\documentclass{article}

     \PassOptionsToPackage{numbers, compress}{natbib}

     \usepackage[final]{neurips_2022}

\usepackage[utf8]{inputenc} 
\usepackage[T1]{fontenc}    
\usepackage{hyperref}       
\usepackage{url}            
\usepackage{booktabs}       
\usepackage{amsfonts}       
\usepackage{nicefrac}       
\usepackage{microtype}      
\usepackage[dvipsnames]{xcolor}         
\usepackage{soul}

\usepackage{caption}
\usepackage{graphicx}
\graphicspath{{./images/}}
\usepackage[normalem]{ulem}
\usepackage{color,hyphenat,balance,booktabs}
\usepackage[fleqn,tbtags]{mathtools}
\usepackage{overpic}
\usepackage{wrapfig}
\usepackage{contour}
\usepackage{enumitem}
\setlist{nosep}
\usepackage{physics}
\usepackage{amsmath}
\usepackage[ruled]{algorithm2e}

\SetCommentSty{mycommfont}
\SetAlgorithmName{List}{List}{List of Lists}

\title{Discovering Design Concepts for CAD Sketches}

\author{%
  Yuezhi Yang\thanks{Work done during internship at Microsoft Research Asia.} \\
  The University of Hong Kong\\
  Microsoft Research Asia \\
  \texttt{yzyang@cs.hku.hk} \\
  \And
  Hao Pan \\
  Microsoft Research Asia \\
  \texttt{haopan@microsoft.com} \\
}

\begin{document}

\maketitle

\begin{abstract}
Sketch design concepts are recurring patterns found in parametric CAD sketches. 
Though rarely explicitly formalized by the CAD designers, these concepts are implicitly used in design for modularity and regularity. 
In this paper, we propose a learning based approach that discovers the modular concepts by induction over raw sketches. 
We propose the dual implicit-explicit representation of concept structures that allows implicit detection and explicit generation, and the separation of structure generation and parameter instantiation for parameterized concept generation, to learn modular concepts by end-to-end training. 
We demonstrate the design concept learning on a large scale CAD sketch dataset and show its applications for design intent interpretation and auto-completion.
\end{abstract}

\section{Introduction}
\label{sec:intro}

\begin{figure}[t]
\begin{minipage}{1.0\linewidth}
	\centering
	\begin{overpic}[width=1.0\linewidth]{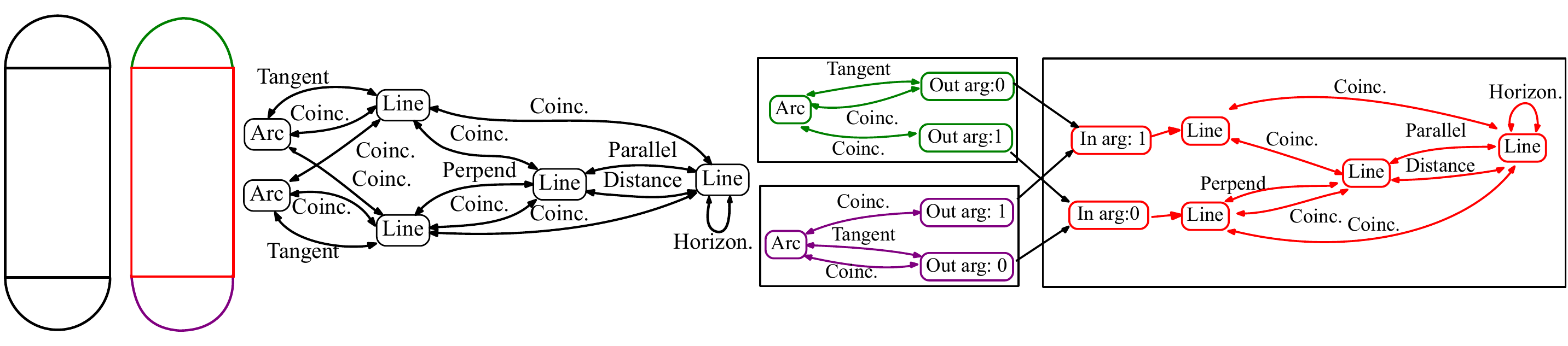}
		\put(0,21.2){\footnotesize (a)}
		\put(53,19.5){\scriptsize $\textcolor{Green}{S.t_0} : \vb{T}_0^1$}
		\put(53,1.5){\scriptsize $\textcolor{Purple}{S.t_1} : \vb{T}_0^1$}
		\put(79,19.5){\scriptsize $\textcolor{Red}{S.t_2} : \vb{T}_1^1$}
	\end{overpic}
\end{minipage}
\begin{minipage}{1.0\linewidth}
\footnotesize
(b) Learned program that restructures the sketch:
\vspace{-2mm}
\scriptsize
$$\begin{aligned}
	\centering
	\vb{T}_0^1 \rightarrow & \lambda(\alpha_0^o, \alpha_1^o).\{t_0{:}\textrm{Arc}, t_1{:}\textrm{Tang.}, t_2,t_3{:}\textrm{Coinc.}, R{=}\{ t_1(t_0, \alpha_0^o), t_2(t_0,\alpha_0^o), t_3(t_0, \alpha_1^o) \} \} \\
	\vb{T}_1^1 \rightarrow & \lambda(\alpha_0^i, \alpha_1^i).\{ t_0{,}t_1{,}t_2{,}t_3{:}\textrm{Line}, t_4{,}t_5{,}t_6{,}t_7{:}\textrm{Coinc.}, t_8{:}\textrm{Perpend.}, t_9{:}\textrm{Parallel}, t_{10}{:}\textrm{Distance}, t_{11}{:}\textrm{Horizon}, \\
	& R{=}\{ t_4(t_0,t_3), t_5(t_0,t_2), t_0(t_1,t_2), t_7(t_1,t_3), t_8(t_1,t_2), t_9(t_2,t_3), t_{10}(t_2,t_3), t_{11}(t_3), \alpha_0^i(t_1), \alpha_1^i(t_0) \} \} \\
	S \rightarrow & \{ \textcolor{Green}{t_0}{,}\textcolor{Purple}{t_1}{:}\vb{T}_0^1, \textcolor{Red}{t_2}{:}\vb{T}_1^1, R{=}\{ t_0(t_2.\alpha_1^i, t_2.\alpha_0^i), t_1(t_2.\alpha_0^i,t_2.\alpha_1^i) \} \}
\end{aligned}$$
\vspace{-3mm}
    \captionof{figure}{\textbf{Concept learning from sketch graphs}. \textbf{(a)} In black are the raw sketch and its constraint graph, with nodes showing primitives and edges depicting constraints.
	Colored are the restructured sketch and its modular constraint graph, where each module box represents a concept; primitives and constraint edges are colored according to the modular concepts. \textbf{(b)} The restructured sketch graph in our DSL program representation (List~{\ref{alg:dsl_syntax_semantics}}), where the whole sketch $S$ is compactly constructed with three instances of two learned $\mathbb{L}^1$ types. We simplify notation super/sub-scripts for readability.} 
\label{fig:teaser}
\end{minipage}
\vspace{-2mm}
\end{figure}

Parametric CAD modeling is a standard paradigm for mechanical CAD design nowadays. In parametric modeling, CAD sketches are fundamental 2D shapes used for various 3D construction operations.
As shown in Fig.~\ref{fig:teaser}, a CAD sketch is made of primitive geometric elements (e.g. lines, arcs, points) which are constrained by different relationships (e.g. coincident, parallel, tangent); the sketch graph of primitive elements and constraints captures design intents, and allows adaptation and reuse of designed parts by changing parameters and updating all related elements automatically \cite{camba2016parametric}.
Designers are therefore tasked with the meticulous design of such sketch graphs, so that the inherent high-level design intents are easy to interpret and disentangle.
To this end, meta-structures (Fig.~\ref{fig:teaser}), which we call \textit{sketch concepts} in this paper, capture repetitive design patterns and regulate the design process with more efficient intent construction and communication \cite{IntentTemplate2021,RevisitDesignIntent2018}.
Concretely, each sketch concept is a structure that encapsulates specific primitive elements and their compositional constraints, and the interactions of its internal elements with outside only go through the interface of the concept.

How to discover these modular concepts automatically from raw sketch graphs?
In this paper, we cast this task as a program library induction problem by formulating a domain specific language (DSL) for sketch generation, where a sketch graph is formalized as a program, and sketch concepts are modular functions that abstract primitive elements and compose the program (Fig.~\ref{fig:teaser}).
Discovering sketch concepts thus becomes the induction of library functions from sketch programs.
While previous works address the general library induction problem via expensive combinatorial search \cite{Houdini2018,DreamCoder2021,Jones2021shapeMOD}, we present a simple end-to-end deep learning solution for sketch concepts.
Specifically, we bridge the implicit and explicit representations of sketch concepts, and separate concept structure generation from parameter instantiation, so that a powerful deep network can detect and generate sketch concepts, by training with the inductive objective of reconstructing sketch with modular concepts.

We conduct experiments on large-scale sketch datasets \cite{SketchGraphs}.
The learned sketch concepts show that they provide modular interpretation of design sketches.
The network can also be trained on incomplete input sketches and learn to auto-complete them. 
Comparisons with state-of-the-art approaches that solve sketch graph generation through autoregressive models show that the modular sketch concepts learned by our approach enable more accurate and interpretable completion results. 

To summarize, we make the following contributions in this paper:
\begin{itemize}
	\item We formulate the task of discovering modular CAD sketch concepts as program library induction for a declarative DSL modeling sketch graphs.
	\item We propose a self-supervised deep learning framework that discovers modular libraries for the DSL with simple end-to-end training.
	\item We show the framework learns from large-scale datasets sketch concepts capturing intuitive and reusable components, and enables structured sketch interpretation and auto-completion.
\end{itemize}

\section{Related work}

\textbf{Concept discovery for CAD sketch } 
It is well acknowledged in the CAD design community that design intents are inherent to and implicitly encoded by the combinations of geometric primitives and constraints \cite{RevisitDesignIntent2018,Kyratzi2022}.
However, there is generally no easy approach to discover the intents and make them explicit, albeit through manual design of meta-templates guided by expert knowledge \cite{IntentTemplate2021,Kyratzi2022}.
We propose an automatic approach to discover such intents, by formulating the intents as modular structures with self-contained references, and learning them through self-supervised inductive training with simple objectives on large raw sketch dataset.
Therefore, we provide an automatic approach for discovering combinatorially complex structures through end-to-end neural network learning.

\textbf{Generative models for CAD sketch } 
A series of recent works \cite{CADAsLanguage2021,Willis2021CVPR,SketchGen2021,seff2022vitruvion,DeepCAD} use autoregressive models \cite{transformerNIPS2017} to generate CAD sketches and constraints modeled through pointer networks \cite{PointerNetwork}.
These works focus on learning from large datasets \cite{SketchGraphs} to generate plausible layouts of geometric primitives and their constraints, which can then be fine-tuned with a constraint solver for more regular sketches.
Different from these works, our aim is to discover modular structures (i.e. sketch concepts) from the concrete sketches. 
Therefore, our framework provides higher-level interpretation of raw sketches and more transparent auto-completion than these works (cf. Sec.~\ref{sec:results}).

\textbf{Program library induction for CAD modeling }  Program library induction has been studied in the shape modeling domain \cite{Jones2021shapeMOD}.
General program synthesis assisted by deep learning is a research topic with increasing popularity \cite{Houdini2018,EllisGraphicsProgram2018,EllisREPL2019,TianShapeProgram2019,DreamCoder2021}.
The library induction task specifically involves combinatorial search, as has been handled by neural guided search \cite{Houdini2018,DreamCoder2021} or by pure stochastic sampling \cite{Jones2021shapeMOD}. 
We instead present an end-to-end learning algorithm for sketch concept induction.
In particular, based on key observations about sketch concepts, we present implicit-explicit dual representations of concept library functions, and separate the concept structure generation from parameter instantiation, to enable self-supervised training with induction objectives.

\section{CAD sketch concept formulation}
\label{sec:definition}

To capture the notion of sketch concepts precisely, we formulate a domain specific language (DSL) (syntax given in List~\ref{alg:dsl_syntax_semantics}, an exhaustive list of data types given in the supplementary).
In the DSL, we first define the basic data types, including \textit{length, angle, coordinate,} and the \textit{reference} type, where a reference binds to another reference or a primitive for modeling the constraint relationships.
Second, we define the $\mathbb{L}^0$ collection of primitive and constraint types as given in raw sketches. 
In particular, we regard the constraints as functions whose arguments are the references to bind with primitives, e.g. a coincident constraint $c = \lambda(r_1,r_2:\textrm{Ref}).\{\}$, where a function is represented in the lambda calculus style (one may refer to \cite{pierce2002types} for introductory lambda calculus formality).
Some constraints have parameters other than mere references, which are treated as variables inside, e.g. parallel distance in List \ref{alg:dsl_syntax_semantics}\footnote{While other works \cite{SketchGen2021,seff2022vitruvion} have skipped such constraints, we preserve them but omit generating the parameter values that can be reliably deduced from primitives. See more discussions in the supplementary.}.
Third, we define the sketch concepts as $\mathbb{L}^1$ types composed of $\mathbb{L}^0$ types.
To be specific, a composite type $\vb{T}_i^1 \in \mathbb{L}^1$ is a function with arguments $[\alpha_k]$ and members $t^0_{i,j}:\vb{T}_{j}^0\in\mathbb{L}^0$, which are connected through a composition operator $R_{\vb{T}^1_i} = \{p(q)|p,q\in[t^0_{i,j}]{\cup}[\alpha_k]\}$ that specifies how each pair of primitive elements binds together. 
For example,  a coincident constraint $p=\lambda(r_1, r_2).\{\}$ may take a line primitive $q$ as its first argument and bind to an argument $\alpha_k$ of the composite type as its second argument, i.e. $p(q,\alpha_k) \in R_{\vb{T}^1_i}$; on the other hand, an argument $\alpha_k$ may bind to a primitive $q$, which is specified by $\alpha_k(q) \in R_{\vb{T}^1_i}$.
Finally, an input sketch $S$ is restructured as a collection of composite types $t^1_i : \vb{T}_i^1\in \mathbb{L}^1$, as well as their connections specified by a corresponding composition operator $R_S$.
$R_S$ records how different concepts bind through their arguments, which further transfers to $\mathbb{L}^0$ typed elements inside the concepts and translates into the raw constraint relationships of the sketch graph.
Fig.~\ref{fig:teaser}(b) shows an example DSL program encoding sketches and concepts.

\begin{algorithm}[tb]
	\tcp{Basic data types}
	Length, Angle, Coord, Ref 
	
    \tcp{$\mathbb{L}^0$ primitive types}
	Line $\rightarrow$ $c_{start\_x}, c_{start\_y}, c_{end\_x}, c_{end\_y}$ : Coord 
	
    Circle $\rightarrow$ $c_{center\_x}, c_{center\_y}$ : Coord,  $l_{radius}$ : Length 
    
    $\cdots$ 
    
    \tcp{$\mathbb{L}^0$ constraint types}
	Coincident $\rightarrow$ $\lambda(r_1, r_2 : \textrm{Ref}).\{\}$ 
	
    Parallel Distance $\rightarrow$ $\lambda(r_1, r_2 : \textrm{Ref})$.\{$l_{dist}$ : Length\} 
    
	$\cdots$ 
	
	\tcp{$\mathbb{L}^1$ composite types}
	$\vb{T}^1_{i}$ $\rightarrow$ $\lambda([ \alpha_k : \textrm{Ref} ]).\{t^0_{i,j}$ : $\vb{T}_{j}^0\in \mathbb{L}^0, R_{\vb{T}_i^1}\left([t^0_{i,j}]{\cup}[\alpha_k]\right)\}$ 
	
	\tcp{Sketch decomposition}
	$S \rightarrow \{ t^1_i : \vb{T}_i^1 \in \mathbb{L}^1, R_S([t^1_i]) \}$ 
	
	\caption{A domain-specific language formulating CAD sketch concepts} 
	\label{alg:dsl_syntax_semantics}
\end{algorithm}

Given the explicit formulation of CAD sketches through a DSL, the discovery of sketch concepts becomes the task of learning program libraries $\mathbb{L}^1$ by induction on many sketch samples.
Therefore, our task resembles shape program synthesis that aims at building modular programs for generating shapes \cite{DreamCoder2021,Jones2021shapeMOD}, and differs from works that use autoregressive language models to generate CAD sketch programs one token at a time \cite{CADAsLanguage2021,SketchGen2021,seff2022vitruvion}.
In Sec.~\ref{subsec:auto_complete}, we show that the structured learning of CAD sketches enables more robust auto-completion than unstructured language modeling.

The search of structured concepts is clearly a combinatorial problem with exponential complexity, which is intractable unless we can exploit the inherent patterns in large-scale sketch datasets.
However, to enable deep learning based detection and search of structured concepts, we need to bridge the implicit deep representations and the explicit and interpretable structures, which we build through the following two key observations:
\begin{itemize}
    \item \textbf{A concept has dual representations}: implicit and explicit. The implicit representation as embeddings in latent spaces is compatible with deep learning, while the explicit representation provides structures on which desired properties (e.g. modularity) can be imposed.
    \item \textbf{A concept is a parameterized structure}. A concept is a composite type with fixed modular structure for interpretability, but the structure is always instantiated by assigning parameters to its component primitives when the concept is found in a sketch.
\end{itemize}

\subsection{Method overview}

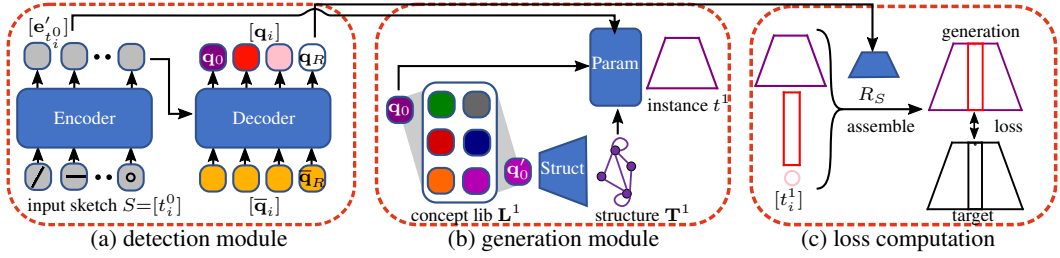
\begin{figure}
    \centering
    \scriptsize
    \begin{overpic}[width=\linewidth]{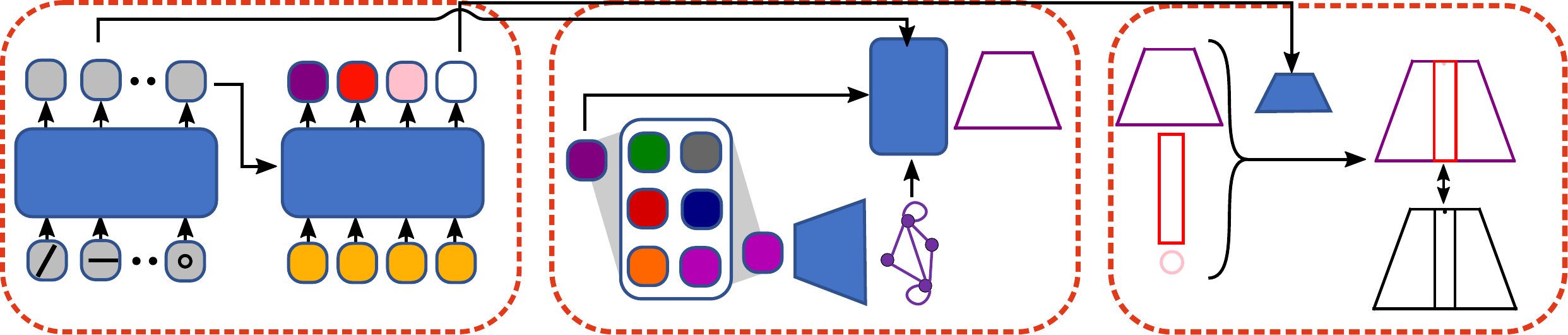}
        \put(2,1.5){input sketch $S{=}[t^0_i]$}
        \put(2,18.5){$[\vb{e}'_{t_i^0}]$}
        \put(23,1.5){$[\overline{\vb{q}}_{i}]$}
        \put(23,18){$[\vb{q}_{i}]$}
        \put(27.8,15.7){$\vb{q}_{R}$}
        \put(27.8,4.2){$\overline{\vb{q}}_{R}$}
        \put(18.6,15.7){\textcolor{white}{$\vb{q}_0$}}        \put(36.3,10.9){\textcolor{white}{$\vb{q}_0$}}
        \put(47.6,5){\textcolor{white}{$\vb{q}_0'$}}
        \put(38.5,0.5){concept lib $\vb{L}^1$}
        \put(56,0.5){structure $\vb{T}^1$}
        \put(61,11){instance $t^1$}
        \put(73,2){$[t_i^1]$}
        \put(81,12){$R_S$}
        \put(80,9){assemble}
        \put(89,18){generation}
        \put(94,9){loss}
        \put(90,0.5){target}
        \put(50.8,5){\textcolor{white}{Struct}}
        \put(55.7,15){\textcolor{white}{Param}}
        \put(4.5,9.8){\textcolor{white}{Encoder}}
        \put(21.5,9.8){\textcolor{white}{Decoder}}
        \put(8,-1.9){\small (a) detection module}
        \put(42,-1.9){\small (b) generation module}
        \put(75.6,-1.9){\small (c) loss computation}
    \end{overpic}
    \vspace{0.3mm}
    \caption{\textbf{Framework illustration}. \textbf{(a)} The detection module is a transformer network that detects from the sketch sequence $[t^0_i]$ implicitly encoded concepts $[\vb{q}_i]$ and their composition $\vb{q}_R$. \textbf{(b)} Each $\vb{q}$ is quantized against the concept library $\vb{L}^1$ to obtain prototype $\vb{q}'$, which is expanded by structure network into an explicit structure $\vb{T}^1$ and further instantiated by parameter network into $t^1$.
    \textbf{(c)} The collection of $[t^1_i]$ are assembled by composition operator $R_S$ generated from $\vb{q}_R$ to obtain the final generated sketch graph, which is compared with the input sketch for loss computation.  }
    \label{fig:pipeline}
    \vspace{-3mm}
\end{figure}

According to the two observations, we design an end-to-end sketch concept learning framework by self-supervised induction on sketch graphs. 
As shown in Fig.~\ref{fig:pipeline}, the framework has two main steps before loss computation: a detection step that generates implicit representations of concepts making up the input sketch, and an explicit generation step that expands the implicit concepts into concrete structures on which self-supervision targets like reconstruction and modularity are applied. 

Building on a state-of-the-art detection architecture \cite{DETR},
the detection module $D$ takes a sketch $S$ as input and detects the modular concepts within it, i.e. $\{\vb{q}_i\} = D(S,\{\overline{\vb{q}}_i\})$, where the concepts are represented implicitly as latent codes $\{\vb{q}_i\}$, and $\{\overline{\vb{q}}_i\}$ are a learnable set of concept instance queries.
Notably, we apply vector quantization to the latent codes and obtain $\{\vb{q}'_i = \min_{\vb{p}\in\mathbb{L}^1}{||\vb{p} - \vb{q}_i||_2}\}$, which ensures that each concept is selected from the common collection of learnable concepts $\mathbb{L}^1$ used for restructuring all sketches.

The explicit generation module is separated into two sub-steps, structure generation and parameter instantiation, which ensures that the modular concept structures are explicit and reused throughout different sketch instances. 
Specifically, the structure network takes each quantized concept code $\vb{q}'_i$ and generates its explicit form $\vb{T}_i^1$ in terms of primitives and constraints of $\mathbb{L}^0$ types along with the composition operator $R_{\vb{T}_i^1}$. 
Subsequently, the parameter network instantiates the concept structure by assigning parameter values to each component of $\vb{T}_i^1$ conditioned on $\vb{q}_i$ and input sketch, to obtain $t^1_i$.

The composition operator $R_S$ for combining $\{t^1_i\}$ is generated from a special latent code $\vb{q}_{R}$ transformed by $D$ from a learnable token $\overline{\vb{q}}_{R}$ appended to $\{\overline{\vb{q}}_{i}\}$. 

The entire model is trained end-to-end by reconstruction and modularity objectives.
In particular, we design loss functions that measure differences between the generated and groundtruth sketch graphs, in terms of both per-element attributes and pairwise references.
Given our explicit modeling of encapsulated structures of the learned concepts, we can further enhance the modularity of the generation by introducing a bias loss that encourages in-concept references.

\section{End-to-end sketch concept induction}
\label{sec:generation}

\subsection{Implicit concept detection}
\label{subsec:backbone}

\textbf{Sketch encoding } 
A raw sketch $S$ can be serialized into a sequence of $\mathbb{L}^0$ primitives and constraints.
Previous works have adopted slightly different schemes to encode the sequence \cite{CADAsLanguage2021,SketchGen2021,seff2022vitruvion,Willis2021CVPR,DeepCAD}.
In this paper, we build on the previous works and take a simple strategy akin to \cite{SketchGen2021,DeepCAD} for input sketch encoding.
Specifically, we split each $\mathbb{L}^0$ typed instance $t^0$ into several tokens: \textit{type}, \textit{parameter}, and a list of \textit{references}.
For each of the token category, we use a specific embedding module.
For example, parameters as scalars are quantized into finite bins before being embedded as vectors (see supplementary for the quantization details), and since there are at most five parameters for each primitive, we pack all parameter embeddings into a single code.
On the other hand, each constraint reference as a primitive index is directly embedded as a code.
Therefore, each token of a $\mathbb{L}^0$ typed instance is encoded as
\begin{equation}
    \vb{e}_{t^0.x} = \textrm{enc}_{type}(t^0) + \textrm{enc}_{pos}(t^0.x) + \left[\textrm{enc}_{param}(t^0.x) | \textrm{enc}_{ref}(t^0.x)\right],
\end{equation}
where $t^0.x$ iterates over the split tokens (i.e., type, parameters, references), the type embedding is shared for all tokens of the instance, the position embedding counts the token index in the whole split-tokenized sequence of $S$, and parameter or reference embeddings are applied where applicable.

\textbf{Concept detection }
We build the detection network as an encoder-decoder transformer following \cite{DETR}. 
The transformer encoder operates on the sketch encoded sequence $[\vb{e}_{t_i^0 \in S}]$ and produces the contextualized sequence $[\vb{e}'_{t_i^0 \in S}]$ through layers of self-attention and feed-forward. 
The transformer decoder takes a learnable set of concept queries $[\overline{\vb{q}}_i]$ of size $k_{qry}$ plus a special query $\overline{\vb{q}}_{R}$ for composition generation, and applies interleaved self-attention, cross-attention to $[\vb{e}'_{t_i^0}]$ and feed-forward layers to obtain the implicit concept codes $[\vb{q}_i]$ and $\vb{q}_R$.
The concept codes are further quantized into $[\vb{q}'_i]$ by selecting concept prototypes from a library $\vb{L}^1$ implicitly encoding $\mathbb{L}^1$, before being expanded into explicit forms.

\subsection{Explicit concept structure generation}
\label{subsec:struct_gen}

\textbf{Concept structure expansion }
Given a library code $\vb{q}'\in \vb{L}^1$ representing a type $\vb{T}^1 \in \mathbb{L}^1$, through an MLP we expand its explicit structure as a collection of codes $[\vb{t}_i^0]$ representing the $\mathbb{L}^0$ type instances $[t_i^0]$ and a matrix representing the composition $R_{\vb{T}^1}$ of $[t_i^0]$ and arguments (cf. List~\ref{alg:dsl_syntax_semantics}).

\begin{wrapfigure}{r}{0.205\textwidth}
  \vspace{-3mm}
  \begin{center}
  \begin{overpic}[width=0.2\textwidth]{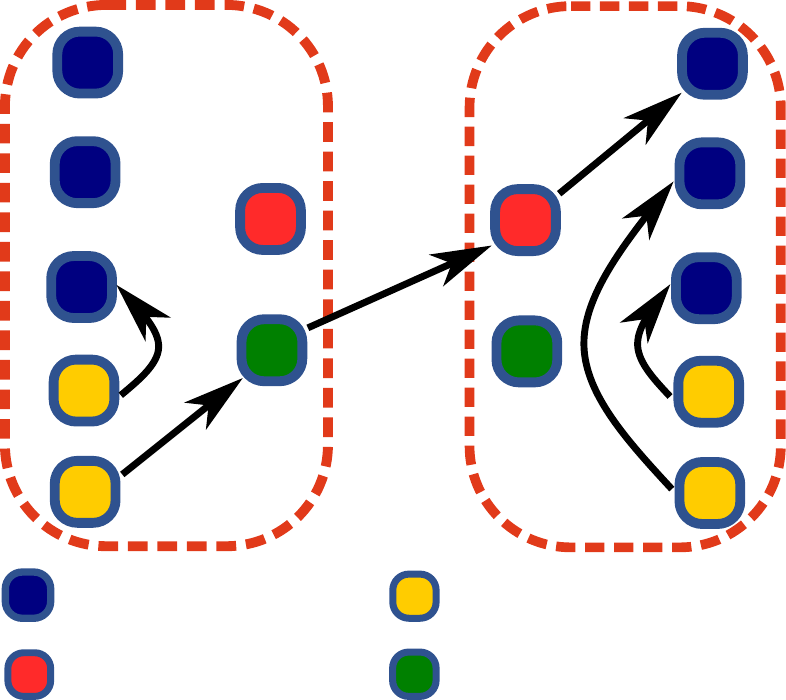}
  \scriptsize
       \put(3,91){concept A}
       \put(61,91){concept B}
       \put(43,90){$R_S$}
       \put(10,10){primitive}
       \put(60,10){constraint}
       \put(10,0){inward arg}
       \put(60,0){outward arg}
  \end{overpic}
  \end{center}
  \vspace{-4mm}
\end{wrapfigure}
We fix the maximum number of $\mathbb{L}^0$ type instances to $k_{L^0}$ (12 by default), and split the arguments into two groups, \textit{inward arguments} and \textit{outward arguments}, each of maximum number $k_{arg}$ (2 by default).
Each type code $\vb{t}_i^0$ is decoded into discrete probabilities over $\mathbb{L}^0$ with an additional probability for null type $\phi$ to indicate the emptiness of this element (cf. Sec.~\ref{subsec:recon_loss}), by $\textrm{dec}_{type}(\cdot)$ as the inverse of $\textrm{enc}_{type}(\cdot)$ in Sec.~\ref{subsec:backbone}.
An inward argument only points to a primitive inside the concept structure and originates from a constraint outside, and conversely an outward argument only points to primitives outside and originates from a constraint inside the concept (see inset for illustration); the split into two groups eases composition computation, as discussed below.

The composition operator $R_{\vb{T}^1}$ is implemented as an assignment matrix $\vb{R}_{\vb{T}^1}$ of shape $(2k_{L^0} {+} k_{arg})\times(k_{L^0} {+} k_{arg})$, where each row corresponds to a constraint reference or inward argument, and each column to a primitive or outward argument.
The two-fold coefficient of constraint references comes from that any constraint we considered in the dataset \cite{SketchGraphs} has at most two arguments.
Each row is a discrete probability distribution such that $\sum_j{\vb{R}_{\vb{T}^1}[i,j]} = 1$, with the maximum entry signifying that the $i$-th constraint/outward argument refers to the $j$-th primitive/inward argument.
We compute $\vb{R}_{\vb{T}^1}$ by first mapping the concept code $\vb{q}'$ to a matrix of logits in the shape of $\vb{R}_{\vb{T}^1}$, and then applying softmax transform for each row.
Notably, we avoid the meaningless loops of an element referring back to itself, and inward arguments referring to outward arguments, by masking the diagonal blocks $\vb{R}_{\vb{T}^1}[2i{:}2i{+}2,i], i {\in} [k_{L^0}]$ and the argument block $\vb{R}_{\vb{T}^1}[2k_{L^0}{:},k_{L^0}{:}]$ by setting their logits to $-\infty$.

\textbf{Cross-concept composition }
Aside from references inside a concept, references across concepts are generated to complete the whole sketch graph.
We achieve cross-concept references by argument passing (see inset above for illustration).
In particular, we implement the cross-concept composition operator $R_S$ as an assignment matrix $\vb{R}_S$ of shape $(k_{qry}{\cdot}k_{arg})\times (k_{qry}{\cdot}k_{arg})$ directly mapped from $\vb{q}_R$ through an MLP.
Similar to the in-concept composition matrix, each row of the cross-concept matrix is a discrete distribution such that $\sum_{j}{\vb{R}_S[i,j]} = 1$, with the maximum entry signifying that the $(i\bmod k_{arg})$-th outward argument of the $\lfloor i/k_{arg} \rfloor$-th concept instance refers to the $(j\bmod k_{arg})$-th inward argument of the $\lfloor j/k_{arg} \rfloor$-th concept instance.

The complete cross-concept reference is therefore the product of three transport matrices:
\begin{equation}
    \centering
    \vb{R}_{cref}[t_i^1,t_j^1] = \vb{R}_{t^1_i}[{:}2k_{L^0},k_{L^0}{:}] \times \vb{R}_{S}[i{\cdot}k_{arg}{:}(i+1){\cdot}k_{arg}, j{\cdot}k_{arg}{:}(j+1){\cdot}k_{arg}] \times \vb{R}_{t^1_j}[2k_{L^0}{:},{:}k_{L^0}],
\end{equation}
where $\vb{R}_{cref}[t_i^1,t_j^1]$ is a block assignment matrix of shape $2k_{L^0}{\times} k_{L^0}$.
Intuitively, $\vb{R}_{cref}[t_i^1,t_j^1]$ specifies how constraints inside $t_i^1$ refers to primitives of $t_j^1$ throughout all possible paths crossing the arguments of two concepts.

Collectively, we denote the complete reference matrix for all pairs of generated $\mathbb{L}^0$ elements as $\vb{R}$ of shape $(2k_{qry}{\cdot} k_{L^0})\times (k_{qry}{\cdot} k_{L^0})$, which includes in-concept and cross-concept references.

\subsection{Concept instantiation by parameter generation}
\label{subsec:param_gen}

Instantiating a concept structure requires assigning parameters to the components where applicable.
Therefore, as shown in Fig.~\ref{fig:pipeline}, the parameter generation network takes a concept structure $\vb{T}^1$ and its implicit instance encoding $\vb{q}$ as input, and produces the specific parameters for each $\mathbb{L}^0$ typed instances inside the concept.
In addition, as the parameters of generated instances are directly related to the input parameters of the raw sketch $S$, we find it improves convergence and accuracy by allowing the parameter network to attend to the input tokens.

We implement the parameter network as a transformer decoder in a similar way as \cite{DETR}.
The instance code $\vb{q}$ is first expanded to $k_{L^0}$ tokens by a small MLP, which are summed with $[\vb{t}_{i}^0]$ token-wise to obtain the query codes.
The parameter network then transforms the query codes through interleaved layers of self-attention, cross-attention to the contextualized input sequence $[\vb{e}'_{t^0}]$, and feed-forward, before finally mapped to explicit parameters in the form of probabilities over quantized bins, through a decoding layer $\textrm{dec}_{param}(\cdot)$ that is inverse of $\textrm{enc}_{param}(\cdot)$ in Sec.~\ref{subsec:backbone}.

\section{Induction objectives}
\label{sec:objectives}

Without any given labels of concepts, we use the following objectives to supervise the inductive network training: sketch reconstruction, concept quantization, and modularity enhancement.

\subsection{Reconstruction loss}
\label{subsec:recon_loss}

As discussed in Sec.~\ref{sec:generation}, an input sketch is restructured into a set of sketch concepts which are expanded into a graph of primitives and constraints; the generated sketch graph $\widetilde{S}$ is compared with the input sketch $S$ for reconstruction supervision.

The comparison of generated and target graphs requires a one-to-one correspondence between elements of the two graphs, on which the graph differences can be measured. 
However, it is nontrivial to find such a matching, because not only are there variable numbers of elements in the two graphs, but also both \textit{elements} and \textit{references between elements} must be taken into account for matching.
To this end, we build a cost matrix that measures the difference for each pair of generated and target elements, in terms of their attributes and references, and apply linear assignment matching on the cost matrix \cite{Munkres1957,Kuhn1955} to establish the optimal correspondence between two graphs.

\textbf{Cost matrix construction }
To compare each pair of generated element and target element, we measure their type differences, and further use type casting to interpret the generated element as the target type, so that their parameters can be compared.
To account for reference differences between the two elements, we compare the reference arrows by the differences of their pointed primitives.

\begin{wrapfigure}{r}{0.2\textwidth}
  \vspace{-2mm}
  \begin{center}
  \begin{overpic}[width=0.16\textwidth, trim=0 -9mm 0 0]{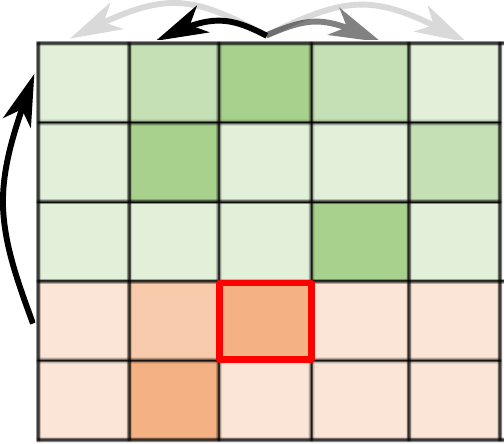}
  \footnotesize
       \put(-18,35){\rotatebox{90}{targets}}
       \put(17,109){generations}
       \put(21,6){cost matrix}
       \put(21,-5){\scriptsize \textcolor{lime}{green}: primitives}
       \put(21,-16){\scriptsize \textcolor{orange}{orange}: constraints}
       \put(-5,36){$p$}
       \put(-12,82){$p.r$}
       \put(47,97){$q$}
       \put(67,97){$q.r{\sim}\vb{R}[2q{+}r,:]$}
  \end{overpic}
  \end{center}
  \caption*{\small Binary cost between target constraint $p$ and generated element $q$.}
  \vspace{-5mm}
\end{wrapfigure}
Specifically, given the target graph $S$ of $k_{tgt}$ elements and the generated graph $\widetilde{S}$ of $k_{qry}{\cdot}k_{L^0}$ elements, we build the cost matrix $\vb{C}$ of shape $k_{tgt}\times(k_{qry}{\cdot}k_{L^0})$ in two steps.
First, for a pair of target element $p$ and generated element $q$, we compare their type and parameter differences by cross-entropy.
We denote the cost matrix in this stage as $\vb{C}_{ury}$, as it accounts for the element-wise unary distances between two graphs.
Second, to measure the binary distances of references, for each target constraint element $p$ and its $r$-th referenced primitive $p.r$, its distance from the generated references of element $q$ is computed as (also illustrated by inset figure):
\begin{equation}
    \vb{C}_{bry}[p,q] = \sum_{r\in\{0,1\}}{ \sum_{j\in \widetilde{S}}{\vb{R}[2q+r,j]\times \vb{C}_{ury}[p.r,j]} },
    \label{eq:binary_cost}
\end{equation}
where $\vb{R}[2q+r,j]$ is the probability of $q$ taking $j$ as its $r$-th reference, as predicted by the composition operation (Sec.~\ref{subsec:struct_gen}).
Intuitively, the binary cost is a summation of unary costs weighted by predicted reference probabilities, where the unary costs measure how different a generated pointed primitive is from the target pointed primitive.
The complete cost matrix is $\vb{C} = w_{ury}\vb{C}_{ury} + w_{bry}\vb{C}_{bry}$, with $w_{ury}=50, w_{bry}=1$; we give a larger weight to the unary costs because meaningful binary costs depend on reliable unary costs in the first place, as evident in Eq.~(\ref{eq:binary_cost}).

\textbf{Matching and reconstruction loss }
Given the cost matrix $\vb{C}$, we apply linear assignment to obtain a matching $\sigma: \widetilde{S} \rightarrow S{\cup}\{\phi\}$ between $\widetilde{S}$ and $S$.
Note that the number of elements of these two graphs can be different, but we have chosen $k_{qry}, k_{L^0}$ such that the generated elements always cover the target elements.
Therefore, $\sigma(q) = p$ assigns a matched generation $q {\in} M {\subset} \widetilde{S}$ to a target $p {\in} S$, but assigns the rest unmatched generations $M' = \widetilde{S}{\setminus} M$ to the empty target $\phi$, i.e. $\sigma(M') = \phi$.
The loss terms for matched generations are simply the corresponding cost terms $\vb{C}[\sigma(q),q], q\in M$; for unmatched generations, we supervise its type to be the empty type $\phi$ and neglect its parameters or references.
We denote the average loss of all generated terms as $\mathcal{L}_{recon}$.

Besides matching cost, we also use an additional reference loss to encourage the generated references to be sharp (i.e., $\vb{R}$ being closer to binary). 
This loss complements the binary costs mentioned above by making sure that even if the generated primitives are similar, a generated constraint only refers to one primitive sharply.

We define the sharp reference loss as
\begin{equation}
    \mathcal{L}_{sharp}  = -\frac{1}{|S_c|}\sum_{p\in S_c,r}{ \log{\vb{R}[2\sigma^{-1}(p)+r, \sigma^{-1}(p.r) ]} },    
\end{equation}
where $p$ iterates over the target constraints $S_c$, $\sigma^{-1}(\cdot)$ is the inverse mapping from target element to generation, and we skip a term if $p.r$ does not exist for constraints with one reference.

\subsection{Concept quantization loss}
\label{subsec:quantize_loss}

Following \cite{VQVAE,VQVAE2}, we optimize the concept code quantization against library $\vb{L}^1$ with:
\begin{equation}
    \mathcal{L}_{vq} = \frac{1}{k_{qry}} \sum_{i\in[k_{qry}]}{||\textrm{sg}(\vb{q}_i) - \vb{q}'_i|| + \beta ||\vb{q}_i - \textrm{sg}(\vb{q}'_i)||} ,
\end{equation}
where $\textrm{sg}(\cdot)$ is the stop gradient operation. 
For training stability, we follow \cite{VQVAE2} and replace the first term with EMA updates of $\vb{q}'\in \vb{L}^1$.
Furthermore, we improve spare code usage by reviving unused code in $\vb{L}^1$ periodically \cite{VQVAE2} (please refer to supplementary for details).

\subsection{Modularity enhancement loss}
\label{subsec:modular_bias_loss}

We look for modular $\mathbb{L}^1$ concepts that have rich and meaningful encapsulated structures, rather than arbitrary groups of $\mathbb{L}^0$ elements that rely on cross-group references to recover the graph structures.
This modularity can be enhanced by limiting the use of arguments for sketch concepts.
Instead of allocating very few arguments as a hard constraint, we introduce a soft bias loss to encourage the restrictive use of arguments, which may still cover cases when more arguments are needed for accurate reconstruction.
To be specific, we penalize the accumulated probability of elements pointing to outward arguments:
\begin{equation}
    \mathcal{L}_{bias} = \frac{1}{|S_c|}\sum_{p\in S_c,r}{\sum_{i\in[k_{arg}]}{\vb{R}_{\vb{T}^1\ni \sigma^{-1}(p)}[2\sigma^{-1}(p)+r,i+k_{L^0}]}},
\end{equation}
where again $p$ iterates over the target constraints $S_c$, $\sigma^{-1}(\cdot)$ is the inverse mapping from target element to generation, and $i+k_{L^0}$ slices the reference probabilities to outward arguments.

\subsection{Total loss}

The training objective sums up losses of reconstruction, concept quantization and modularity bias:
\begin{equation}
    \mathcal{L}_{total} =  w_{recon}\mathcal{L}_{recon} + w_{sharp}\mathcal{L}_{sharp} + w_{vq}\mathcal{L}_{vq} + w_{bias}\mathcal{L}_{bias},
\end{equation}
where we empirically use weights $w_{recon}=1,w_{sharp}=20,w_{vq}= 1, w_{bias}= 25$ throughout all experiments unless otherwise specified in the ablation studies.

\section{Results}
\label{sec:results}

\textbf{Dataset and implementation }
Following previous works \cite{CADAsLanguage2021,SketchGen2021,seff2022vitruvion}, we adopt the SketchGraphs dataset \cite{SketchGraphs} which contains millions of real-world CAD sketches for training and evaluation.
We filter the data by removing trivially simple sketches and duplicates, and limit the sketch complexity such that the number of primitives and constraints is within $[20,50]$. As a result, we obtain around 1 million sketches and randomly split them into 950k for training and 50k for testing.
We defer network details to the supplementary and open-source code and data to facilitate future research\footnote{URL to code and data: \url{https://github.com/yyuezhi/SketchConcept}}.

\textbf{Evaluation metrics }
We evaluate the generated sketches in terms of reconstruction accuracy and sketch concept modularity, which are the two major objectives of our task.
We measure reconstruction accuracy by the F-scores of generated primitives and constraints, where F-score is simply the harmonic mean of precision and recall.
A generated primitive is considered a correct match with ground-truth if its type and parameters are correct, where for the scalar parameters we allow a threshold of $10\%$ of quantization levels.
A constraint is correct if and only if its type, parameter and references match ground-truth, i.e., the generated $q$ is correct w.r.t target $p$ iff $q$ has the same type and parameters with $p$ and the primitives $q.r$ and $p.r$ are correctly matched.
Modularity is measured by the percentage of constraints with references entirely within the encapsulating concepts, among all correct constraints.

\begin{minipage}{1.0\linewidth}
    \centering
    \includegraphics[width=1.0\linewidth]{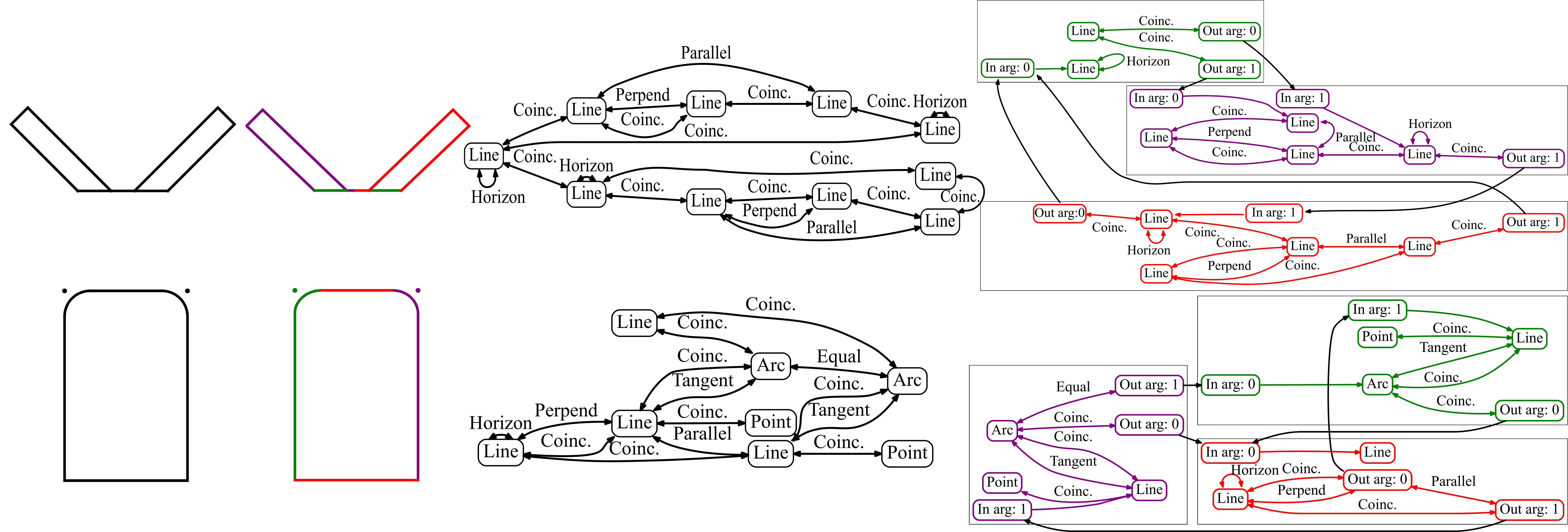}
\vspace{-5mm}
\captionof{figure}{\textbf{Design intent parsing.} Left: input raw sketches and sketches restructured with concepts. Right: raw constraint graphs and modular constraint graphs. Primitives and constraints in the restructured sketches and graphs are colored according to concepts.}
\label{fig:parsing}
\end{minipage}

\begin{minipage}{1.0\linewidth}
    \centering
    \includegraphics[width=1.0\linewidth]{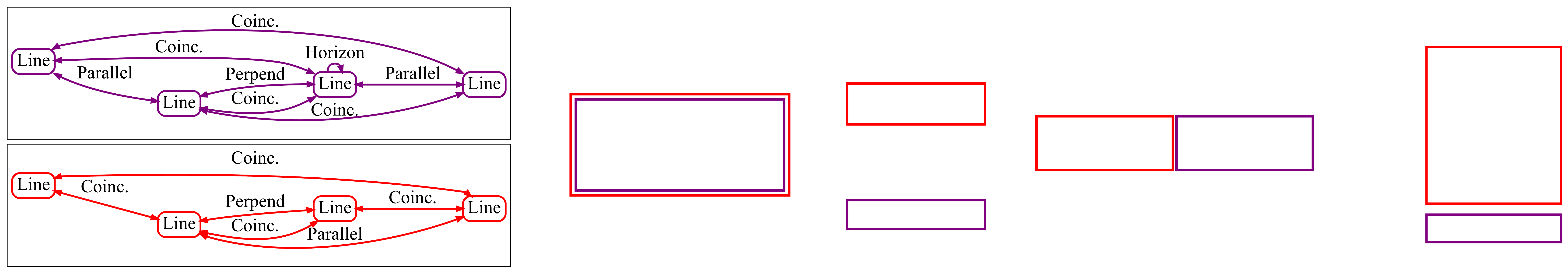}
\vspace{-6mm}
\captionof{figure}{\textbf{Instances of concepts.} Left: two learned rectangle concepts with subtly different structures. Right: four sketches containing instances of these two concepts. }
\label{fig:library_examples}
\end{minipage}

\begin{minipage}{1.0\linewidth}
    \centering
    \begin{overpic}[width=1.0\linewidth]{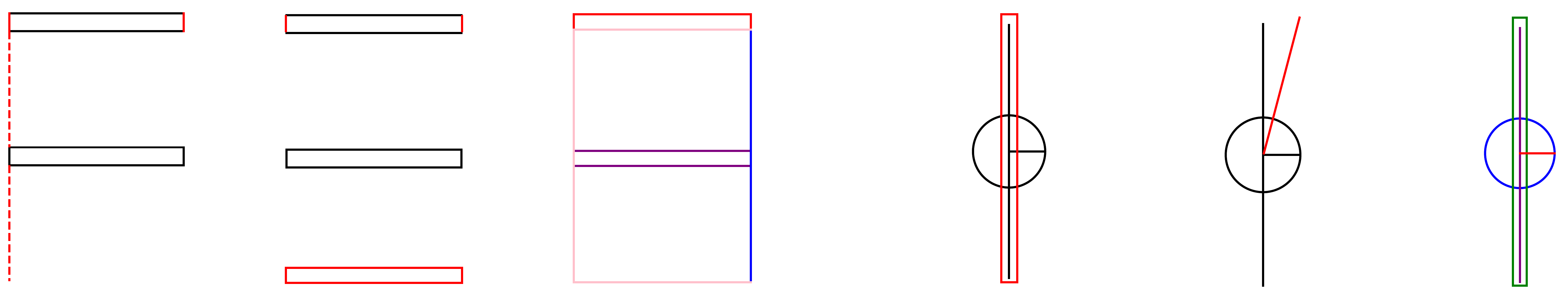}
     \put(1,1){\small (a)}
     \put(60,1){\small (b)}
    \end{overpic}
\vspace{-6mm}
\captionof{figure}{\textbf{Auto completion.} Each example shows the input partial sketch (black) and groundtruth completion (red), result of the autoregressive baseline, and our result (colored by concepts). }
\label{fig:auto_completion}
\end{minipage}

\begin{minipage}{0.49\linewidth}
\centering
\begin{overpic}[width=\linewidth, trim=0 -5mm 0 0]{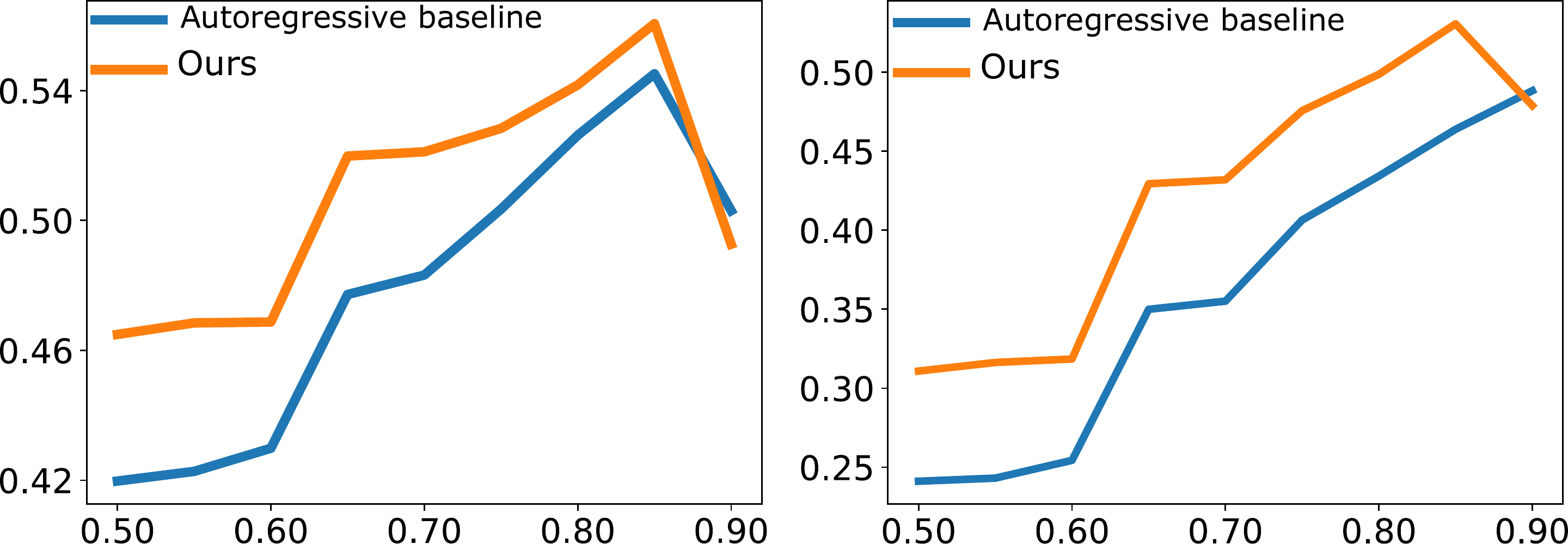}
 \put(15,-2.5){\small (a) primitive}
 \put(65,-2.5){\small (b) constraint}
\end{overpic}
\captionof{figure}{\textbf{Auto completion comparison.} Plotted are F-scores at different ratios of partial input.}
\label{fig:auto_complete_numerics}
\end{minipage}
\hspace{0.5pt}
\begin{minipage}{0.49\linewidth}
\centering
\setlength{\tabcolsep}{2pt} 
{\small
\begin{tabular}{cccc}
    \toprule
    Config & Primitive  & Constraint  & Modular(\%)\\
    \midrule
    $-\vb{C}_{bry}$ & 0.993& 0.290 & 34.2 \\
    $-\mathcal{L}_{sharp}$& 0.983&  0.104& 100 \\
    $-\mathcal{L}_{bias}$ & 0.992 & 0.743 & 13.3\\
    Ours &  0.994 & 0.766 & 50.8 \\
    \bottomrule
\end{tabular}    
}
\captionof{table}{\textbf{Loss ablation.} F-scores are reported for primitives and constraints.}
\label{tab:ablation}
\end{minipage}

\subsection{Design intent interpretation}

By training our model on the raw sketches with self-supervised induction losses, we obtain a result library of sketch concepts and a model for design intent parsing that interprets a given sketch into modular concepts and their combination.
Indeed, we find the automatically discovered concepts capture natural design intents and modular structures.
For example, through the restructured sketches and constraint graphs in Figs.~\ref{fig:teaser} and \ref{fig:parsing}, we find that our network decomposes sketches into modular structures like rectangles, line-arcs and parallel lines that align symmetrically, even though no such prior knowledge is applied during training except for concept modularity.
Fig.~\ref{fig:library_examples} shows that a given concept can be used repetitively in different sketches, and structures with subtle differences in constraint relations can be detected and distinguished into different concepts of the library. Note that these subtle structural differences are subsumed in the input sketch graph, which makes them more difficult to detect. 
We refer to the supplementary for more examples of design intent parsing and instantiation of learned concepts, as well as quantitative analysis of the learned library.

\subsection{Auto completion}
\label{subsec:auto_complete}

Auto-completion is a critical feature of CAD modeling software for assisting designers.
Given a partial sketch of primitives and their constraints, auto-completion aims at complementing them with the rest primitives and constraints to form regular and well-structured designs.
Therefore, our concept detection and generation approach would naturally enhance the auto-completion task with better regularity.
For training and evaluation, following previous work \cite{seff2022vitruvion}, we synthesize the partial input by removing a suffix of random length (up to $50\%$) from the sketch sequence, along with constraints that refer to the removed primitives, and make the model learn to generate the full sketches.

State-of-the-art methods \cite{CADAsLanguage2021,SketchGen2021,seff2022vitruvion,Willis2021CVPR} formulate auto-completion through a combination of primitive and constraint generation models, both of which operate in an autoregressive fashion, with the constraint model conditioned on and referring (by pointers \cite{PointerNetwork}) to the generated primitives. 
Since these works use diverse sketch encodings and have no publicly released code at submission time, for fair comparison, we implement the autoregressive baseline with our sketch encoding (Sec.~\ref{subsec:backbone}).

Fig.~\ref{fig:auto_complete_numerics} compares our method with autoregressive baseline under various primitive mask ratios: our method has superior primitive and constraint accuracy than the autoregressive baseline at almost all mask ratios. 
This difference confirms that since our method completes sketches concept-by-concept instead of primitive-by-primitive, more meaningful structures are likely to be generated.
Our model also gains advantage by taking primitives and constraints together as input and generating primitives and constraints simultaneously, while in comparison the autoregressive baseline separates generation in two steps (primitives followed by constraints). 
Indeed, in practice CAD designers rarely finish all primitives first before supplementing the constraints, but rather apply constraints on partial primitives immediately whenever they form a design intent.
Fig.~\ref{fig:auto_completion} shows how our approach interprets the partial inputs and completes with modular concepts (see supplementary for more examples); in comparison, the autoregressive baseline does not provide such interpretable or regular completions.

\subsection{Ablation study}
\label{subsec:ablation}

To evaluate the impact of different loss terms of the induction objective (Sec.~\ref{sec:objectives}), we train several models in the absence of these losses respectively on the auto-encoding task.
The results are shown in Table.~\ref{tab:ablation}.
We see that removing the binary costs $\vb{C}_{bry}$ from reconstruction loss results in significant drop of constraint reconstruction, showing its necessity for constraint reference learning.
Removing sharp reference loss $\mathcal{L}_{sharp}$ similarly fails constraint reference learning, although modularity enhancement bias loss makes all constraint references inside concepts.
Removing the modularity enhancement bias loss $\mathcal{L}_{bias}$ only results in a slight drop in reconstruction quality but a significant drop in modularity, since without it cross-concept reference through arguments is more likely and therefore modularity suffers.
We provide more ablation tests on hyper parameters like the numbers of concept queries $k_{qry}$ and arguments $k_{arg}$ in the supplementary.

\section{Conclusion}
\label{sec:conclusion}

CAD sketch concepts are meta-structures containing primitives and constraints that define modular sub-graphs and capture design intents.
By formulating the sketch concepts as program libraries of a DSL, we present an end-to-end approach for discovering CAD sketch concepts through library induction learning.
Key to our approach are the implicit-explicit representation of concepts and the separated structure generation and parameter instantiation for concept generation, which together enable the end-to-end training under self-supervised induction objectives.
By training on large-scale sketch dataset, our approach enables the discovery of repetitive and modular concepts from raw sketches, and more structured and interpretable auto-completion than baseline autoregressive models.

\textbf{Limitations and future work }
Design intents can be hierarchical \cite{Kyratzi2022}, meaning that higher order meta-structures can be built out of lower order ones.
In this sense, our framework only addresses the first order library induction, and should be extended for higher order library learning; toward this goal, we believe a progressive approach like \cite{DreamCoder2021} can be used with our framework as the one-step induction.
In addition, similar strategies of end-to-end induction learning can be applied to constraint graphs involving 3D CAD operations  or even more general programs in other domains, as long as they have similar declarative and parametric structures as sketch graphs.

\bibliographystyle{plainnat}
\bibliography{sketchconcept}

\newpage

\appendix
\section{Supplementary for ``Discovering Design Concepts for CAD Sketches''}

\subsection[The complete list of L0 types]{The complete list of $\mathbb{L}^0$ types}

We provide the complete list of $\mathbb{L}^0$ types in List~\ref{alg:all_l0_types}. 
These types are constructed based on the given data types from the SketchGraphs dataset \cite{SketchGraphs}.
Note that in the current implementation we do not distinguish sub-primitive references that point to different parts of a primitive, but rely on the predicted geometric closeness of primitive parts to tell them in the post-process, as we find the geometric predictions are generally quite accurate for this purpose.
On the other hand, we note that the extension of references into primitive parts can be trivially achieved by turning primitives into functions and augmenting them with arguments (similar to how we model constraints), such that each argument corresponds to a primitive part; the constraint references can then pinpoint to primitive parts through argument passing (Sec.~\ref{subsec:struct_gen}).

\begin{algorithm}
\caption{The complete list of $\mathbb{L}^0$ types considered in this work.}
\label{alg:all_l0_types}
	\tcp{Basic data types}
	Construction, Length, Angle, Coord, Ref
    
    \tcp{$\mathbb{L}^0$ primitive types}
	Line $\rightarrow$ $b_{dash}$: Construction, $c_{start\_x}, c_{start\_y}, c_{end\_x}, c_{end\_y}$ : Coord 
	
    Circle $\rightarrow$ $b_{dash}$: Construction, $c_{center\_x}, c_{center\_y}$ : Coord,  $l_{radius}$ : Length 
    
    Point $\rightarrow$ $b_{dash}$: Construction, $c_{x}, c_{y}$ : Coord
   
   Arc $\rightarrow$ $b_{dash}$: Construction, $c_{center\_x}, c_{center\_y}$ : Coord,  $l_{radius}$ : Length ,  $a_{start}$, $a_{end}$ : Angle
    
    \tcp{$\mathbb{L}^0$ constraint types}
	Coincident $\rightarrow$ $\lambda(r_1, r_2 : \textrm{Ref}).\{\}$ 
	
    Distance $\rightarrow$ $\lambda(r_1, r_2 : \textrm{Ref})$.\{$l_{dist}$ : Length\} 
    
    Horizontal  $\rightarrow$ $\lambda(r_1 : \textrm{Ref}).\{\}$ 
    
	Parallel $\rightarrow$ $\lambda(r_1, r_2 : \textrm{Ref}).\{\}$ 
     
    Vertical $\rightarrow$ $\lambda(r_1 : \textrm{Ref}).\{\}$ 
    
    Tangent $\rightarrow$ $\lambda(r_1, r_2 : \textrm{Ref}).\{\}$ 
    
    Length $\rightarrow$ $\lambda(r_1 : \textrm{Ref})$.\{$l_{dist}$ : Length\} 
    
    Perpendicular $\rightarrow$ $\lambda(r_1, r_2 : \textrm{Ref}).\{\}$ 
    
    Equal $\rightarrow$ $\lambda(r_1, r_2 : \textrm{Ref}).\{\}$ 
    
    Diameter $\rightarrow$ $\lambda(r_1: \textrm{Ref})$.\{$l_{dist}$ : Length\} 
    
    Radius $\rightarrow$ $\lambda(r_1: \textrm{Ref})$.\{$l_{dist}$ : Length\} 
    
    Angle $\rightarrow$ $\lambda(r_1, r_2: \textrm{Ref})$.\{$a_{ang}$ : Angle\} 
    
    Concentric $\rightarrow$ $\lambda(r_1, r_2 : \textrm{Ref}).\{\}$ 
    
    Normal $\rightarrow$ $\lambda(r_1, r_2 : \textrm{Ref}).\{\}$ 
\end{algorithm}

\subsection{Implementation details}
\label{subsec:implementation}

\textbf{Sketch encoding format }In Sec.~\ref{sec:generation} we described how sketches are encoded to allow network learning; here we present more implementation details.

We encode the input sketch $S$ as a series of primitive tokens followed by a series of constraint tokens, with these tokens supplemented by learned positional encoding according to their indices in this sequence (Sec.~\ref{subsec:backbone}).
We additionally insert learnable \texttt{START}, \texttt{END} and \texttt{NEW} tokens at the front of the sequence, the end of the sequence, as well as between every encoded primitive/constraint respectively, to produce the complete sequence. 

\begin{figure}
    \centering
    \includegraphics[width=1.0\linewidth]{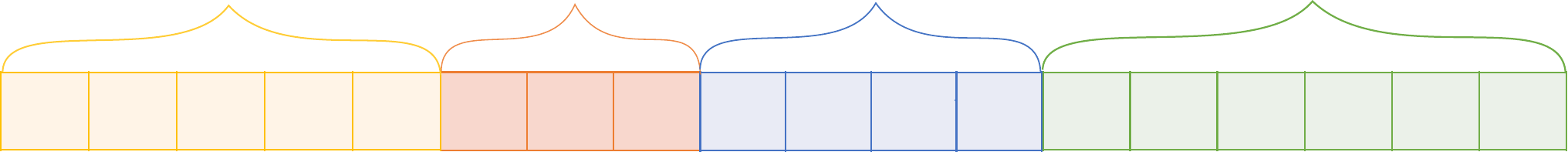}
    \footnotesize
       \put(-347,40){Line}
       \put(-262,40){Point}
       \put(-186,40){Circle}
       \put(-72,40){Arc}
       \put(-396,9){$b_{dash}$}
       \put(-370,9){$c_{sx}$}
       \put(-348,9){$c_{sy}$}
       \put(-325,9){$c_{ex}$}
       \put(-302,9){$c_{ey}$}
       \put(-285,9){$b_{dash}$}
       \put(-258,9){$c_x$}
       \put(-235,9){$c_y$}
       \put(-219,9){$b_{dash}$}
       \put(-192,9){$c_x$}
       \put(-170,9){$c_y$}
       \put(-152,9){$l_{rad}$}
       \put(-132,9){$b_{dash}$}
       \put(-103,9){$c_x$}
       \put(-81,9){$c_y$}
       \put(-64,9){$l_{rad}$}
       \put(-44,9){$a_{start}$}
       \put(-20,9){$a_{end}$}
    \caption{\textbf{Parameter code schema}. A parameter code contains 18 tokens, each of 14 dims, that are concatenated and zero-padded to 256 dims. 
    For a particular primitive type, only tokens corresponding to the specified type are used in the parameter code, the rest tokens are reset to zero. 
    We did not allocate slots for constraint parameters in the current implementation; in comparison, previous works \cite{SketchGen2021,seff2022vitruvion} simply omit constraints with parameters.}

    \label{fig:parameter_schema}
\end{figure}

Each primitive is represented by two consecutive tokens: a $\mathbb{L}^0$ type token and a parameter token. 
The $\mathbb{L}^0$ type of primitive is encoded by a 256-dim embedding, obtained by an embedding layer denoted as $\textrm{enc}_{type}$. 
The parameters of a primitive are encoded in the parameter token; compared with using different numbers of tokens for different primitive types that previous autoregressive baselines do \cite{CADAsLanguage2021,SketchGen2021,seff2022vitruvion}, our one-token parameter encoding allows straightforward matching with a target primitive even if the predicted primitive type does not match the target, which simplifies training.
In particular, we use a schema shown in Fig.~\ref{fig:parameter_schema} to encode the parameter values, where each basic data type is represented by a 14-dim code that is obtained by embedding the quantized parameter value, and all slots of a specific primitive type are used while the rest slots are set zero. 
To represent that the resultant parameter token belongs to a specific primitive type, we augment parameter token with the type token by summing the two vectors to produce the final parameter token (Sec.~\ref{subsec:backbone}). 

Each constraint is represented by a type token and several reference tokens.
The constraint type token is obtained through the same embedding layer $\textrm{enc}_{type}$ as primitives. 
To encode a reference, we use a 256-dim embedding to encode the primitive index in this expanded sequence, obtained through the embedding layer $\textrm{enc}_{ref}$.
We omit constraint parameters in the current implementation, and defer their inference to post-processing according to the positions of predicted primitives;
in comparison, most previous works \cite{SketchGen2021,seff2022vitruvion} have simply skipped constraint types with parameters.

\textbf{Sketch parameter decoding }
$\textrm{dec}_{param}()$ has a mirrored structure of $\textrm{enc}_{param}()$. It takes a latent parameter code as input and decodes it into a 256-dim code (Fig.~\ref{fig:parameter_schema}), which contains several segments corresponding to different primitive types. 
When doing type casting (Sec.~\ref{sec:objectives}), the segment corresponding to the target type is taken for parameter decoding.
Each primitive property is represented by a 14-dim embedding code, from which a quantized property value is recovered by an inverse-embedding layer; during this inverse-embedding process, the logits are processed by argmax to query the quantized value. 
Following previous works \cite{CADAsLanguage2021,SketchGen2021,seff2022vitruvion}, we always work with quantized attribute values as categorical variables during network training and inference.

\textbf{Normalization, augmentation and quantization }We normalize all sketches inside a $2\times2$ square centered at the origin, and remove duplicated sketches by rasterizing into $128\times128$ binary valued images and removing those with the same images. 
We apply random shrinking augmentation with scaling factors of $0.5\sim 0.8$.
The continuous basic data types (List~\ref{alg:all_l0_types}) are uniformly quantized; in particular, we assign 30 bins for \textit{angle}, 20 bins for \textit{length} and 80 bins for \textit{coordinate}.

\textbf{Network and training details }
The detection network is a transformer encoder-decoder network, with the encoder/decoder having 12 layers, 8 attention heads and latent dimension of 256. 

The structure generation network takes a library code $\vb{q}'\in \vb{L}^1$ and generates the $\mathbb{L}^0$ type elements $[t_i^0]$ within and a matrix representing the composition $R_{\vb{T}^1}$ of $[t_i^0]$ and arguments. 
Specifically, the 256-dim library code $\vb{q}'$ first passes through an MLP\footnote{Unless otherwise specified, all MLPs used in this paper have uniform hidden dimensions as the input dimension and ReLU activation after each hidden linear layer.} of 3 layers to expand to $k_{L^0} \times 256$ dims, i.e. $k_{L^0}$ codes representing $[t_i^0]$ elements. 
Then each code passes through another MLP of 3 layers (i.e. $\textrm{dec}_{type}$) to output the discrete probabilities of $\mathbb{L}^0$ types that $t_i^0$ assumes. 
To generate the composition matrix $\vb{R}_{\vb{T}^1}$, we use another MLP of 5 layers to expand the library code to a $(2k_{L^0} {+} k_{arg})\times(k_{L^0} {+} k_{arg})$ matrix and apply softmax on each row, as detailed in Sec.~\ref{subsec:struct_gen}.

The parameter network generates parameters to instantiate concepts. 
It first expands each of the $k_{qry}$ concept instance codes $[\vb{q}_i]$ into $k_{L^0}$ parameter latent codes, which are further added with the corresponding parameter type embeddings obtained from structure generation network and fed into a transformer decoder to generate explicit parameters.
The transformer decoder here has the same hyper parameters as the concept detection decoder (i.e. 12 layers, 8 attention heads, and 256 latent dimension). 
The decoder transforms each group of $k_{L^0}$ latent parameter codes by cross-attending to contextualized input sequences $[\vb{e}'_{t_i^0}]$, and finally maps them to parameter tokens as described in Fig.~\ref{fig:parameter_schema} through $\textrm{dec}_{param}$, which are further decoded into probabilities over quantized basic data types by corresponding inverse embedding layers.

We implement all modules in Pytorch, and use the Adam optimizer with a learning rate of $10^{-4}$ to train the network for 160 epochs on 4 V100 GPUs, which takes 2 days to complete.

\textbf{Library size, EMA code update and dead code revival }
In our experiments, we use a library of $1000$ candidate concepts for $\vb{L}^1$.
We follow \cite{VQVAE2} to replace the first term of concept quantization loss (i.e.  $||\textrm{sg}(\vb{q}_i) - \vb{q}'_i||$ ) with exponential moving average (EMA) updates of $\vb{q}'\in \vb{L}^1$.
Specifically, for each code $\vb{q}_i'$, we define two accumulated variables $n_i \geq 0$ and $\vb{m}_i \in \vb{R}^d$, which are initialized as 1 and a random unit vector, respectively. 
They are later updated in each gradient descent iteration following the rules:
\begin{align}
   n_i &:= \gamma n_i + (1 - \gamma) N_i  \\
   \vb{m}_i &:= \gamma \vb{m}_i + (1 - \gamma) \sum_j{\vb{q}_{i,j}}  \\
   \vb{q}_i' &:= \frac{\vb{m}_i}{n_i}
 \end{align}
where $\{\vb{q}_{i,j}\}$ are $N_i$ detection queries that select $\vb{q}_i'$ as the closet concept prototype in this iteration. 
We set the decay rate $\gamma = 0.99$ and the commitment cost coefficient $\beta = 1$ in all our experiments.

In addition, we find that the concept quantization process may suffer from codebook collapse where all $[\vb{q}_i]$ select to few codes of $\vb{L}^1$, which impairs the capability of the model. 
To solve this problem, in the training process we use dead-code revival \cite{VQVAE2} to periodically (every 100 mini-batches) find an unused code in $\vb{L}^1$ and replace it with the $\vb{q}$ who has farthest distance to its closest code $\vb{q}'$.

\subsection{Autoregressive baseline implementation detail}

Following \cite{Willis2021CVPR,SketchGen2021,seff2022vitruvion}, the autoregressive baseline contains two modules, the primitive model that generates primitives sequentially and the constraint model that takes primitives as input and generates constraints sequentially. 
The primitive model is an autoregressive transformer decoder of 12 layers, 8 attention heads and latent dimension 256. 
The constraint model is a transformer encoder-decoder, where the encoder contextualizes input primitives, and the decoder is an autoregressive model generating constraints.
Constraint reference to primitives is implemented by computing dot product correlation between the generated reference token and contextualized primitive tokens produced by the encoder, following the Pointer Network design \cite{PointerNetwork}.
The constraint model encoder/decoder have the same hyper-parameters as the primitive model.

\subsection{More results}
\begin{figure}
    \centering
    \includegraphics[width=1.0\linewidth]{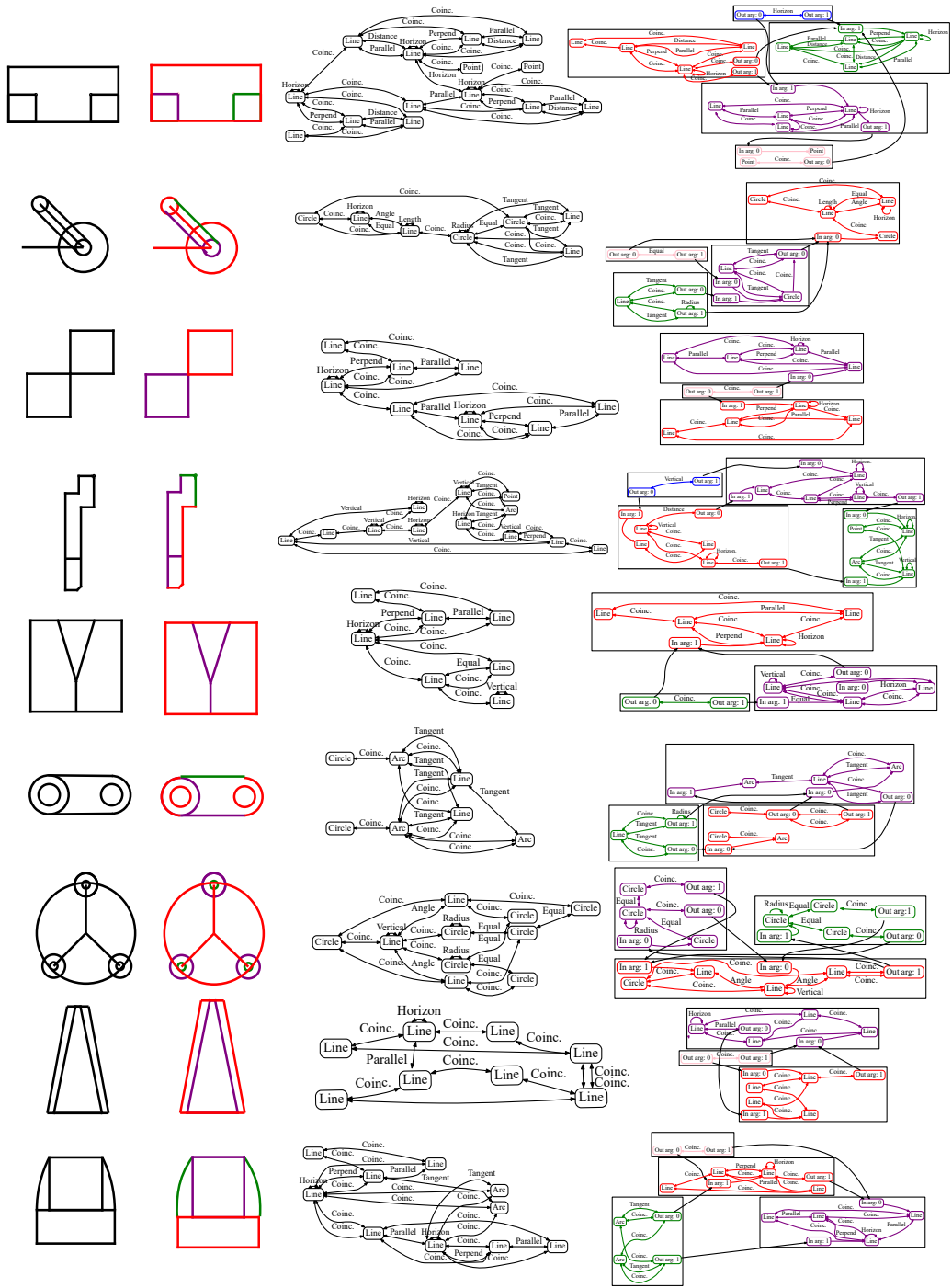}
    \vspace{-15mm}
    \caption{\textbf{Design intent parsing}. Each example shows the input raw sketch and corresponding constraint graph (in black), as well as our interpreted sketch made of modular concepts and corresponding modular constraint graph, where primitives and constraints are colored according to their encapsulating concepts. }
    \label{fig:design_intent_more_results}
\end{figure}

\begin{figure}
    \centering
    \includegraphics[width=1.0\linewidth]{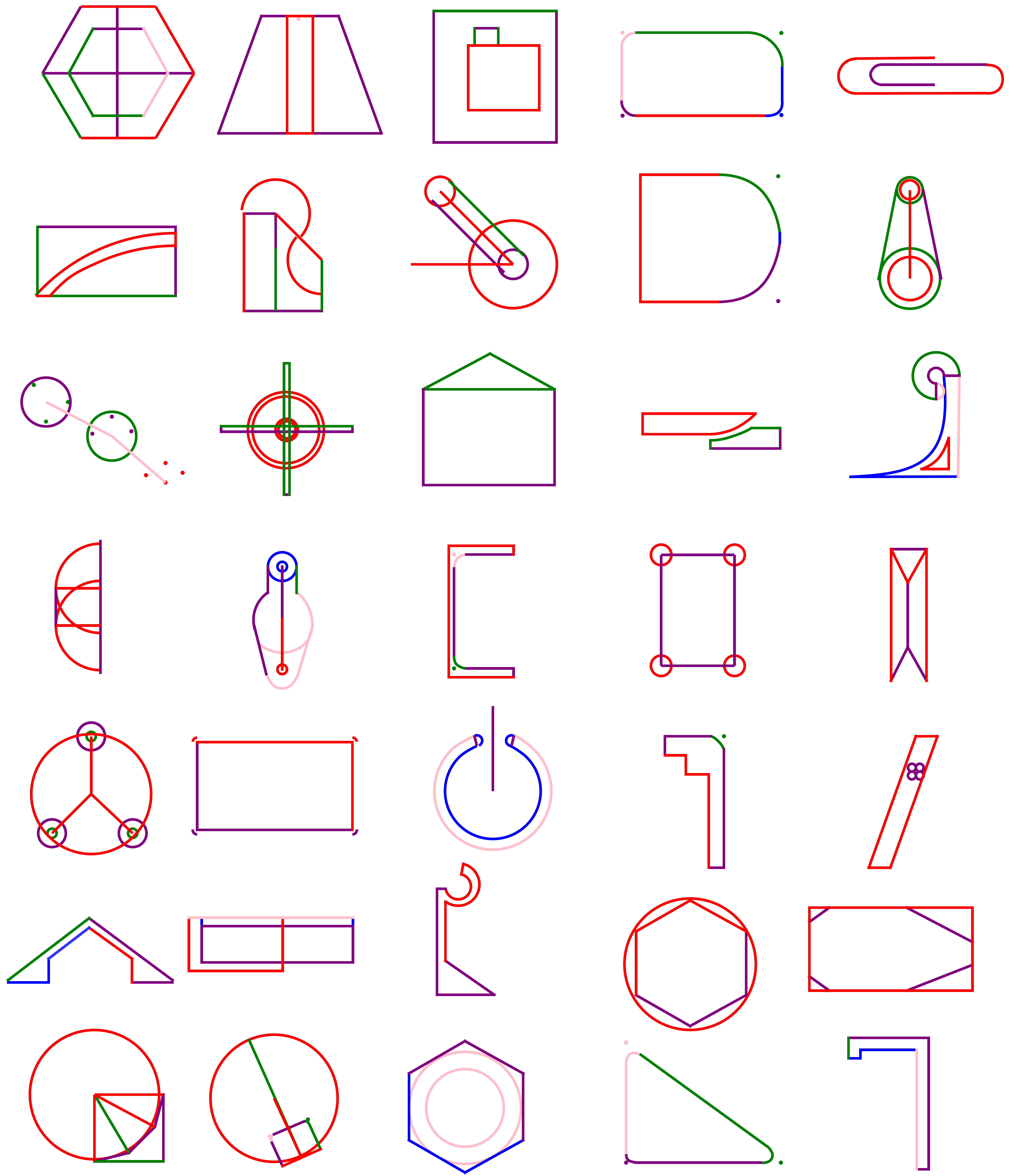}
    \caption{\textbf{Design intent parsing} without showing the constraint graphs. In each sketch example, $\mathbb{L}^0$ primitives of the same color belong to the same sketch concept. }
    \label{fig:design_intent_more_results_no_graph}
\end{figure}

\begin{figure}
    \centering
    \includegraphics[width=0.90\linewidth]{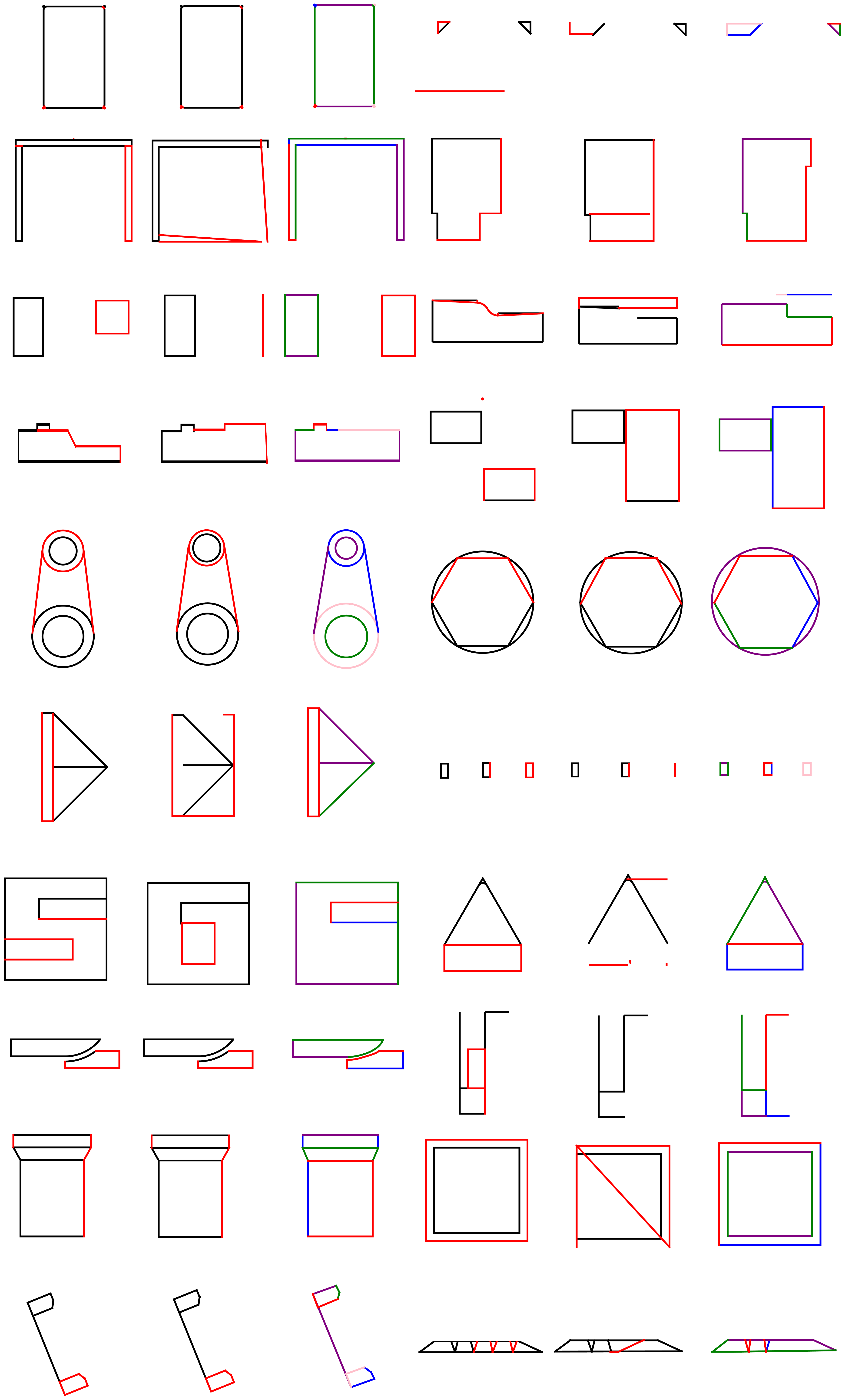}
    \caption{\textbf{More results of auto-completion}. Each example shows the input partial sketch (black) and groundtruth completion (red), result of the autoregressive baseline, and our result (colored by concepts). Our completion results show better interpretability and regularity.}
    \label{fig:auto_complete_more_results}
\end{figure}

We present more results on design intent interpretation and auto completion. 
In Fig.~\ref{fig:design_intent_more_results}, we show more results of how raw sketches are parsed with learned concepts, where primitives are colored according to their encapsulating concepts, and constraint graphs are visualized to show the modular concepts.
In Fig.~\ref{fig:design_intent_more_results_no_graph} we show more such design intent interpretation results without constraint graphs.
Fig.~\ref{fig:auto_complete_more_results} presents more auto-completion results, where again we compare with baseline autoregressive approach and demonstrate better interpretability and more regular completions.

\subsection{Concept library analysis}

\begin{figure}
    \centering
    \begin{overpic}[width=\linewidth]{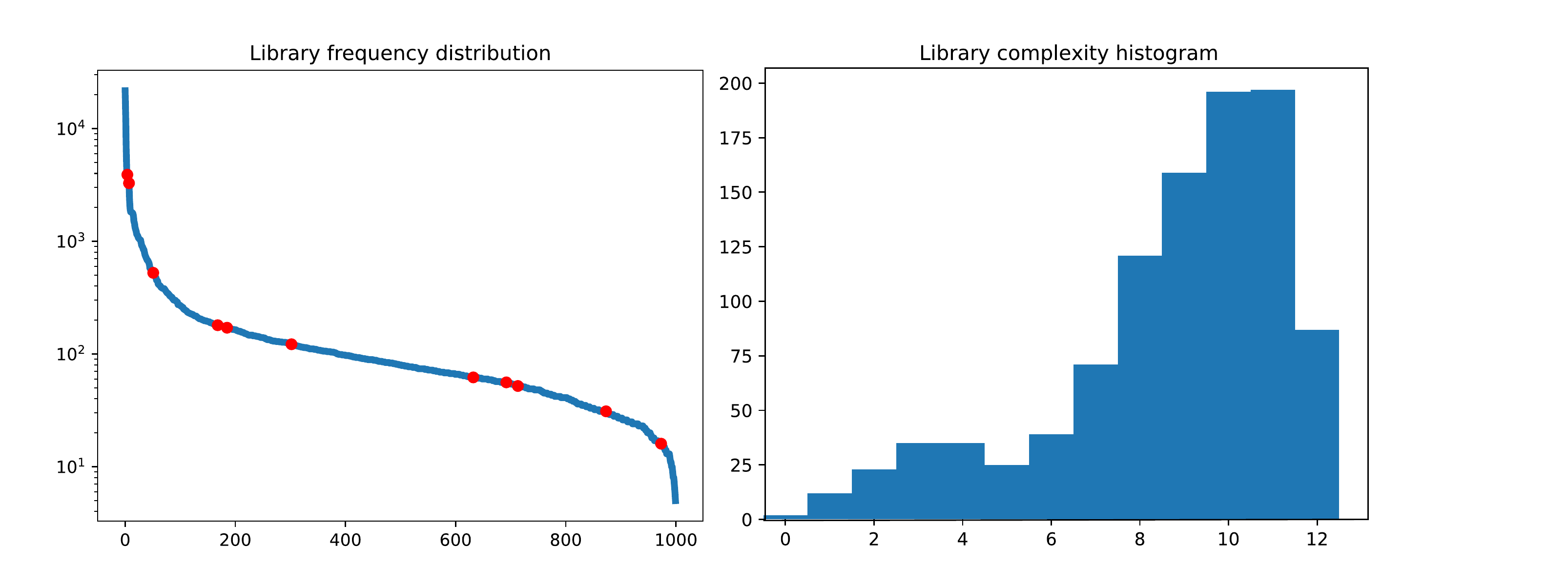}
		\put(24,0){(a)}
		\put(67,0){(b)}
	\end{overpic}
    \caption{ \textbf{Statistics about the learned library concepts}. \textbf{(a)} Library frequency distribution. The horizontal axis shows the 1000 $\mathbb{L}^1$ library concepts learned and sorted according to their frequencies in the test dataset. The vertical axis shows the frequency value in log-scale. The majority of learned concepts have stable but not very high occurrence frequencies, meaning they follow a long-tail distribution as expected. Concepts denoted by red points on the curve are visualized in Fig.~\ref{fig:concept_examples}. \textbf{(b)} Library complexity histogram. The horizontal axis is the number of $\mathbb{L}^0$ instances contained in a concept, and the vertical axis is the number of concepts of a specific complexity. We can see that degenerate (i.e. empty) or trivial (i.e. size 1) concepts are rare among the whole learned library.} 
    \label{fig:lib_dist}
\end{figure}

\begin{figure}
    \centering
    \includegraphics[width=1.0\linewidth]{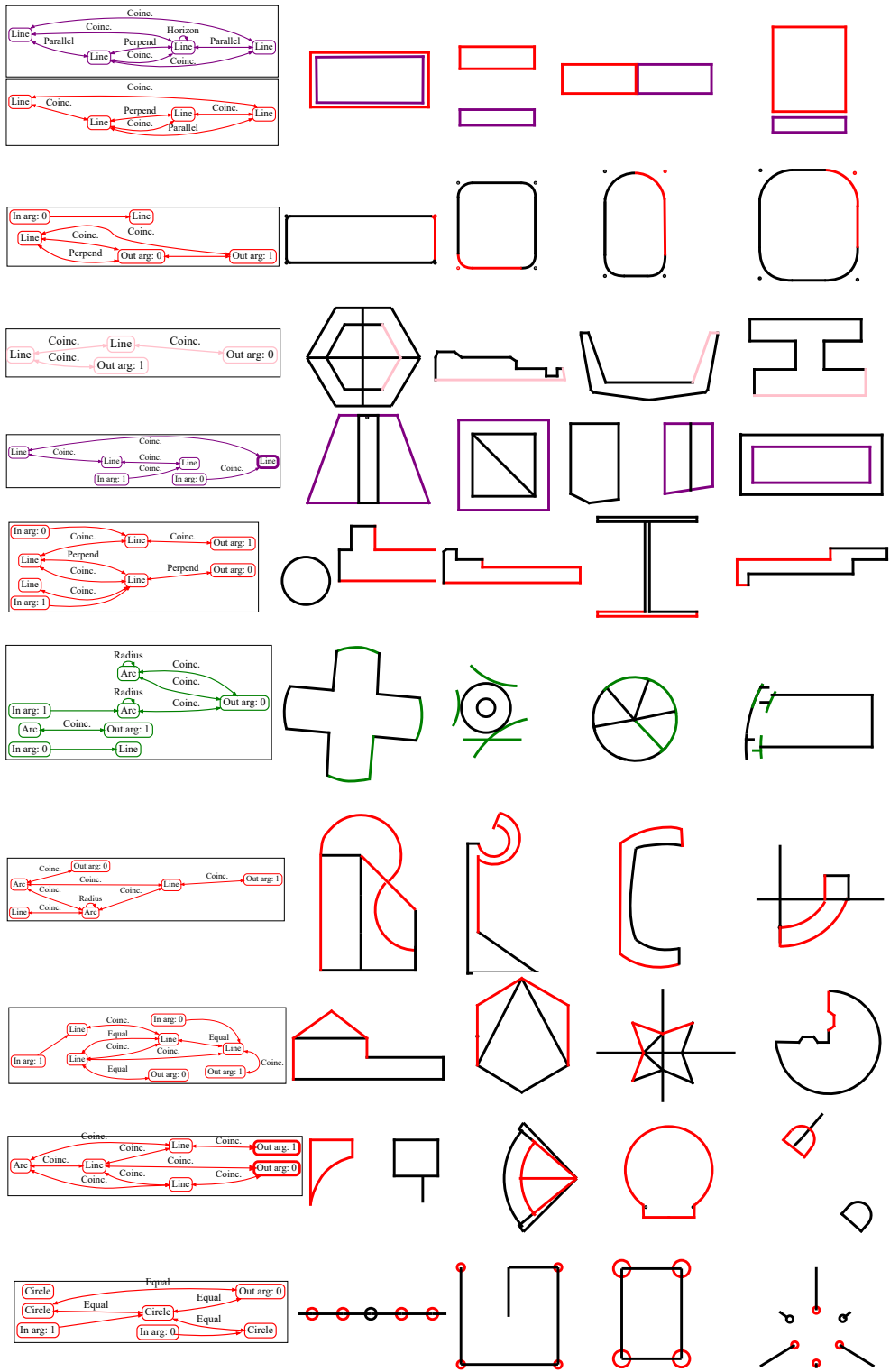}
    \caption{\textbf{Examples of learned library concepts and their corresponding sketch instances}. The concepts are sorted by their occurrence frequency (high to low) in the test set, and correspond to the red dots marked on the distribution curve of Fig.~\ref{fig:lib_dist}.}
    \label{fig:concept_examples}
\end{figure}

Fig.~\ref{fig:lib_dist}(a) shows the frequency of how often our learned library concepts are used in the test dataset.
The distribution shows a long-tail pattern, which is expected because the diversity of sketches demands a wealth of modular sketch concepts that individually may not be used extensively.
We provide more concrete concepts and corresponding sketches containing these concepts in Fig.~\ref{fig:concept_examples}. 
These concepts are arranged according to appearance frequency (from high to low) as marked with red points in Fig.~\ref{fig:lib_dist}(a).
The most frequently used concepts are rectangles with different constraint variants due to their high abundance within regular sketches. 
Besides, concepts with simple structures, e.g. few lines connected together by coincidence, are generally more frequently used than those with complex structures, as the simple structures are more flexible and can fit in diverse sketches. 

Fig.~{\ref{fig:lib_dist}}(b) shows the complexity of learned library concepts in terms of how many $\mathbb{L}^0$ instances are contained in a concept. We can see that there are a small number of degenerate concepts with empty $\mathbb{L}^0$ instances and trivial concepts with only one $\mathbb{L}^0$ instances. The empty concepts serve as placeholder for filling up the gaps between small sketches and the maximal graph of $k_{qry}$ concepts. The trivial concepts exist because we always convert a raw sketch into a set of $\mathbb{L}^1$ concepts, and for those $\mathbb{L}^0$ elements of the raw sketch that do not fit into any modular concept, they will be encapsulated with such trivial $\mathbb{L}^1$ concepts for the sake of complete reconstruction.

\subsection{Parameter refinement with constraint solver}
\begin{figure}
    \centering
    \includegraphics[width=0.8\linewidth]{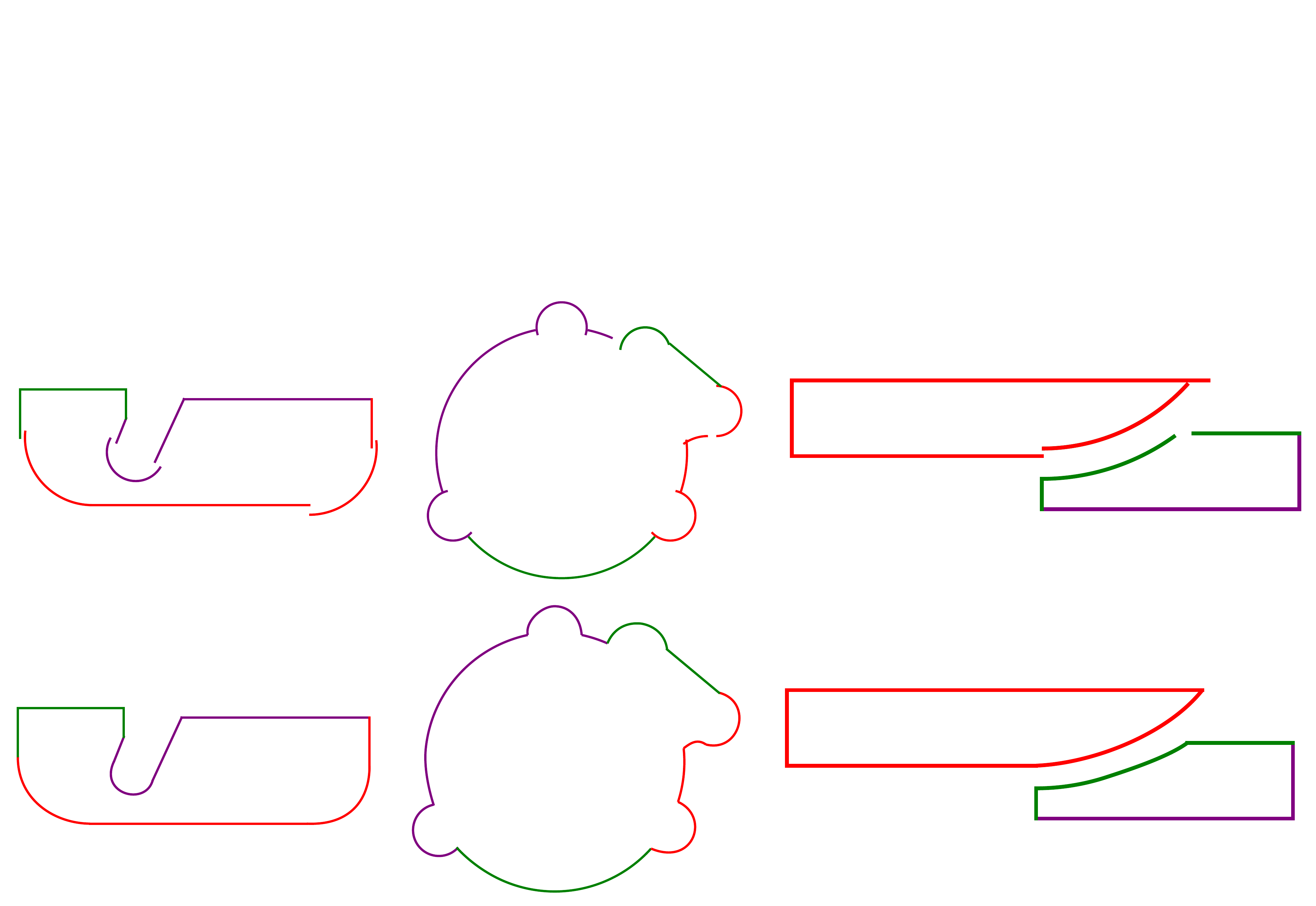}
    \caption{\textbf{Parameter refinement with constraint solver}. Generated sketches have parameter inaccuracies (upper row); constraint solver refines the sketches by applying generated constraints and fitting the primitives together properly (lower row). }
    \label{fig:constraint_solver}
\end{figure}

The errors in generated primitive parameters (e.g. due to quantization of basic data types) can be mitigated by applying constraints with a constraint solver provided by OnShape \cite{onshape}.  In Fig.~\ref{fig:constraint_solver}, we show examples of sketches before and after refining primitive parameters with constraint solver. 

\subsection{More ablation results}

\begin{minipage}{0.49\linewidth}
\centering
\setlength{\tabcolsep}{2pt} 
{\small
\begin{tabular}{cccc}
    \toprule
    $k_{qry}$ & Primitive  & Constraint  & Modular(\%)\\
    \midrule
    5 & 0.994 & 0.766 & 50.8 \\
    6 & 0.994 &  0.808 & 36.0 \\
    8 & 0.991 & 0.845 & 32.6 \\
   10 &  0.998 & 0.894 & 15.9 \\
   12 &  0.999 & 0.918 & 14.9 \\
    \bottomrule
\end{tabular}    
}
\captionof{table}{\textbf{Query number $k_{qry}$ ablation.} F-scores are reported for primitives and constraints.}
\label{tab:ablation_query_num}
\end{minipage}
\hspace{0.01\linewidth}
\begin{minipage}{0.49\linewidth}
\centering
\setlength{\tabcolsep}{2pt} 
{\small
\begin{tabular}{cccc}
    \toprule
    $k_{arg}$ & Primitive  & Constraint  & Modular(\%)\\
    \midrule
    1 & 0.993& 0.666 & 52.6 \\
    2 &  0.993&  0.766 & 50.8 \\
    3 & 0.990 & 0.7577 & 18.5 \\
    4 &  0.994 & 0.776 & 7.1 \\
    \bottomrule
\end{tabular}    
}
\captionof{table}{\textbf{Argument number $k_{arg}$ ablation.} F-scores are reported for primitives and constraints.}
\label{tab:ablation_arg_num}
\end{minipage}

\begin{minipage}{1.0\linewidth}
\centering
\setlength{\tabcolsep}{2pt} 
\begin{tabular}{cccc}
        \toprule
        $\mathbb{L}^1$ size & Primitive  & Constraint  & Modular(\%)\\
        \midrule
        50 & 0.970 &  0.581  & 48.1\\
        100 & 0.980 & 0.600  & 48.8\\
        500 & 0.989 & 0.735  & 49.2\\
        1000 & 0.994 & 0.766  & 50.8\\
        2000 & 0.995 & 0.779  & 50.6\\
        \bottomrule
    \end{tabular}

\captionof{table}{\textbf{$\mathbb{L}^1$ library size ablation.} F-scores are reported for primitives and constraints.}
\label{tab:lib_size}
\end{minipage}

\begin{figure}
    \centering
    \begin{overpic}[width=0.8\linewidth]{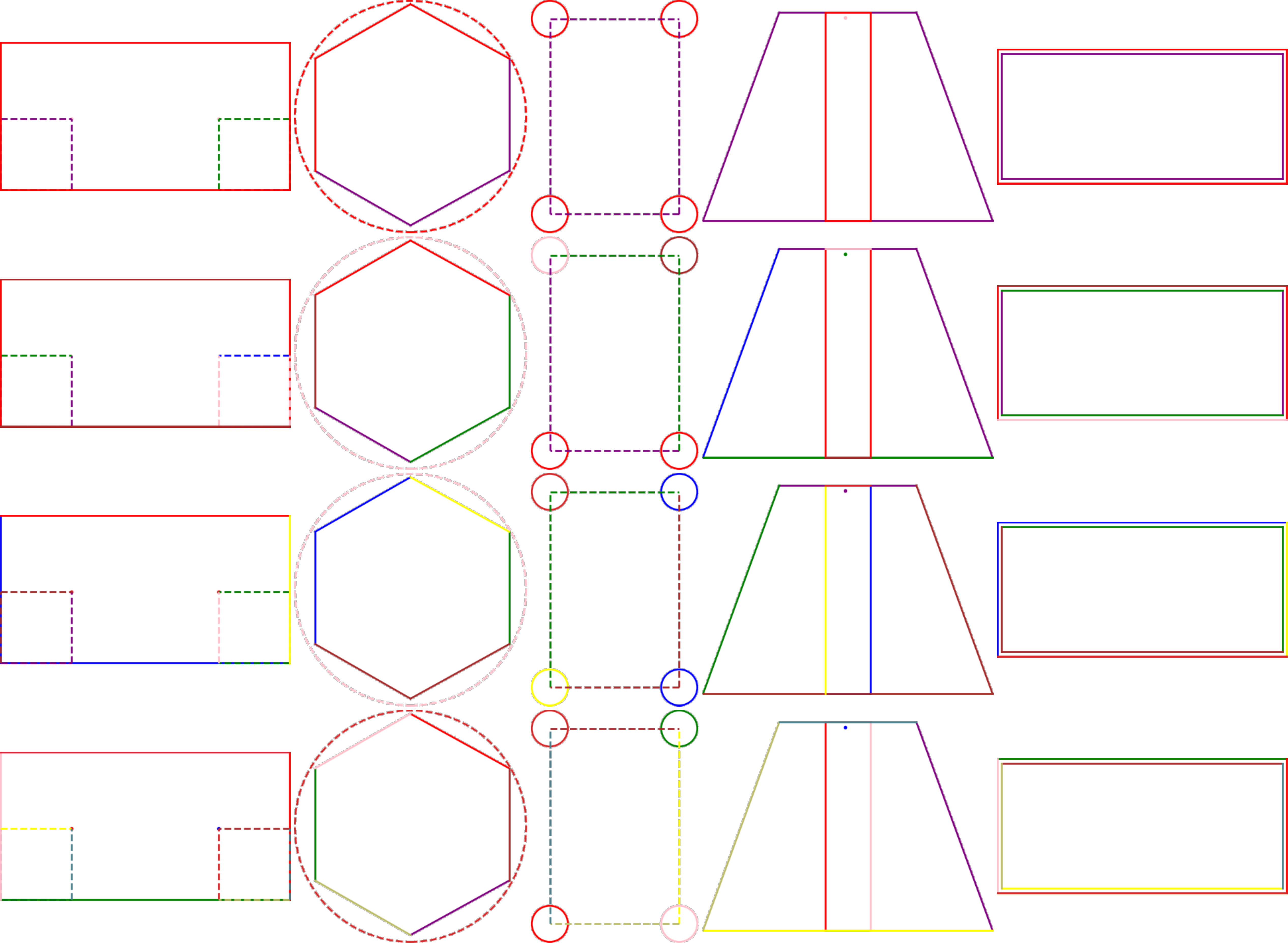}
     \put(-14,8){$k_{qry}= 10$}
     \put(-14,26){$k_{qry}= 8$}
     \put(-14,44){$k_{qry}= 6$}
     \put(-14,63){$k_{qry}= 5$}
    \end{overpic}

    \caption{\textbf{Design intent interpretation trained with different concept query numbers $k_{qry}$}. Larger $k_{qry}$ leads to less modular concepts. }
    \label{fig:struct_vis_query_num}
\end{figure}

\begin{figure}
    \centering
    \begin{overpic}[width=0.8\linewidth]{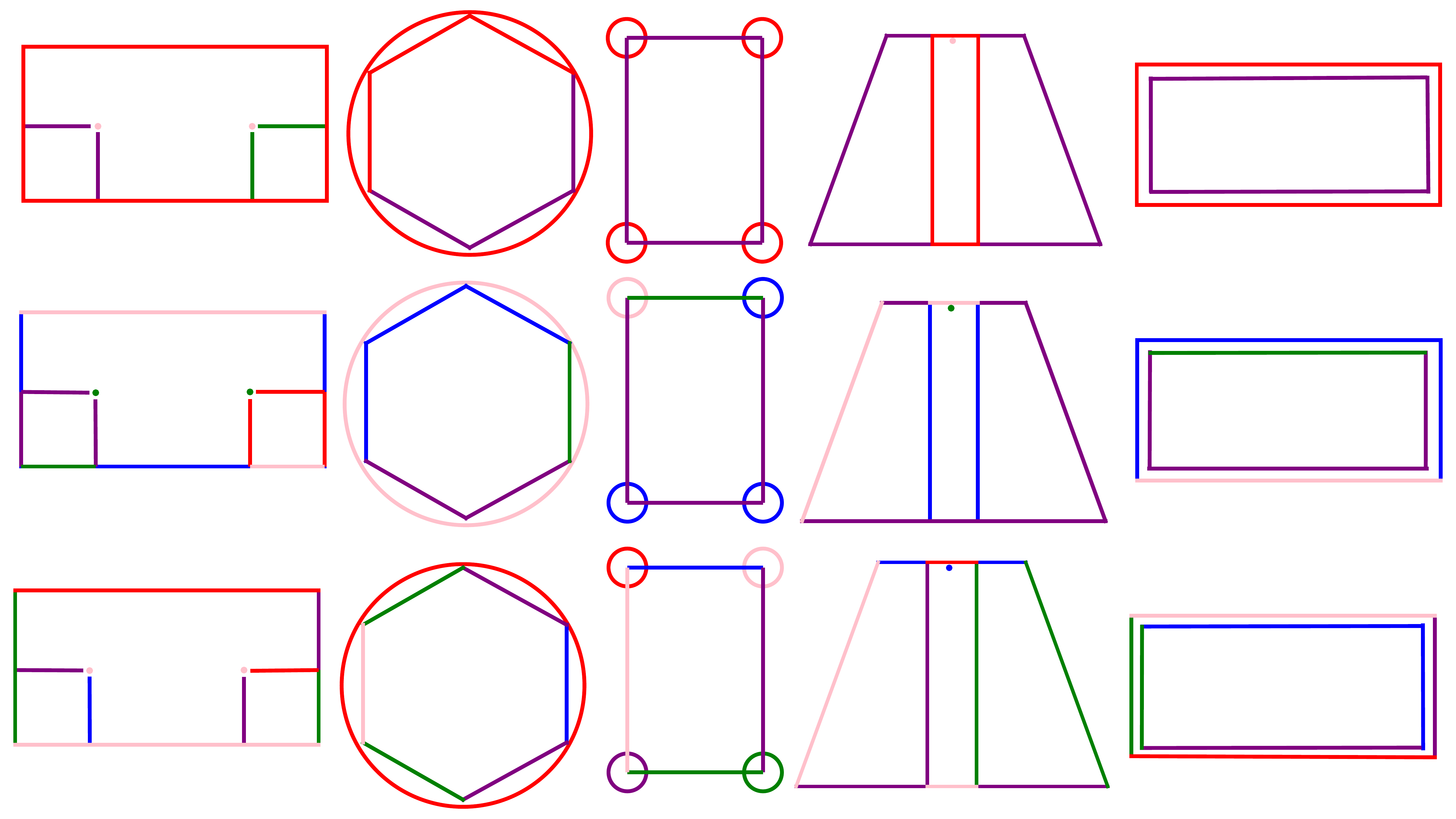}
     \put(-14,11){$k_{arg}=4$}
     \put(-14,30){$k_{arg}=3$}
     \put(-14,47){$k_{arg}=2$}
    \end{overpic}

    \caption{\textbf{Design intent interpretation trained with different argument numbers $k_{arg}$}. Larger $k_{arg}$ leads to less modular concepts. }
    \label{fig:struct_vis_arg_num}
\end{figure}

To evaluate the impact of hyper parameters such as the number of concept queries $k_{qry}$, arguments $k_{arg}$ and library size $\mathbb{L}^1$, we train our model under different $k_{qry}$, $k_{arg}$ and $\mathbb{L}^1$ sizes on the auto-encoding design intent interpretation task. 

When changing $k_{qry}$, we also adjust $k_{L^0}$ the number of $\mathbb{L}^0$ elements each concept contains, so that the total number of generated $\mathbb{L}^0$ elements, i.e. $k_{qry} \times k_{L^0}$, is unchanged.
The evaluation results on adjusting $k_{qry}$, $k_{arg}$, $\mathbb{L}^1$ are shown in Table~\ref{tab:ablation_query_num}, Table~\ref{tab:ablation_arg_num} and Table~\ref{tab:lib_size} respectively.

From Table~\ref{tab:ablation_query_num} we can see that adding more concept queries makes the model more expressive and flexible, demonstrated as the increasing constraint F-scores;
on the other hand, this comes with a cost of hurting modularity, as a sketch can be decomposed into more granular components. 

From Table~\ref{tab:ablation_arg_num} we see that adding more arguments $k_{arg}$ than default 2 does not result in a significant improvement in constraint F-score, but leads to a significant decrease in modularity, suggesting the current default argument number is sufficient. 
On the other hand, decreasing the number of arguments leads to a significant drop in constraint F-score but does not achieve obviously higher modularity. 

We visualize examples of design intent interpretation for different $k_{qry}$ and $k_{arg}$ in Fig.~\ref{fig:struct_vis_query_num} and  Fig.~\ref{fig:struct_vis_arg_num} respectively, to give an intuitive sense of the above numeric results especially on modularity. 

From Table~{\ref{tab:lib_size}} we can see that adding more $\mathbb{L}^1$ library makes the model more expressive and flexible, showing increasing constraint F-scores. However, such improvement becomes marginal when more libraries are introduced, as the newly introduced libraries are mainly used to capture structures that rarely appear and have little impact on the overall results (intuitively, they mainly continue the fall-off trend of far-right tail regions of the frequency distribution shown in Fig.~{\ref{fig:lib_dist}}(a)). On the other hand, the modularity maintains at roughly the same level throughout the changes over library size.

\subsection{Image-conditioned generation}

We extend our model to image-conditioned generation, where we are interested in accurately recovering a parametric sketch from an image of hand-drawn sketch. 
For comparison, we also extend the auto-regressive baseline to this image-conditioned generation. 

Following \cite{SketchGen2021,seff2022vitruvion}, we use a ViT style encoder to condition the generation on images. 
Specifically, the input sketch image of size $128\times 128$ is partitioned into non-overlapping square patches of size $16\times 16$. 
The image patches are flattened and pass through an MLP of 3 layers to produce a sequence of 64 image tokens (each of dimension 256), and then feed into a transformer encoder to produce contextualized image embeddings that the detection decoder cross-attends to. 
For autoregressive baseline, we similarly augment the primitive model with such an image encoder and use cross-attention to image tokens in the autoregressive primitive decoder.
The ViT style image encoder used here has the same hyper parameters as the other transformer modules discussed above (i.e. 12 layers, 8 attention heads, 256 latent dimension). 

We train our model with learning rate of $3\times 10^{-4}$ for 200 epochs and the autoregressive baseline with the same learning rate for 400 epoch to convergence. 
We used the xkcd packages in mathplotlib to simulate sketches of hand-drawn style. 

\begin{figure}
    \centering
    \includegraphics[width=1.0\linewidth]{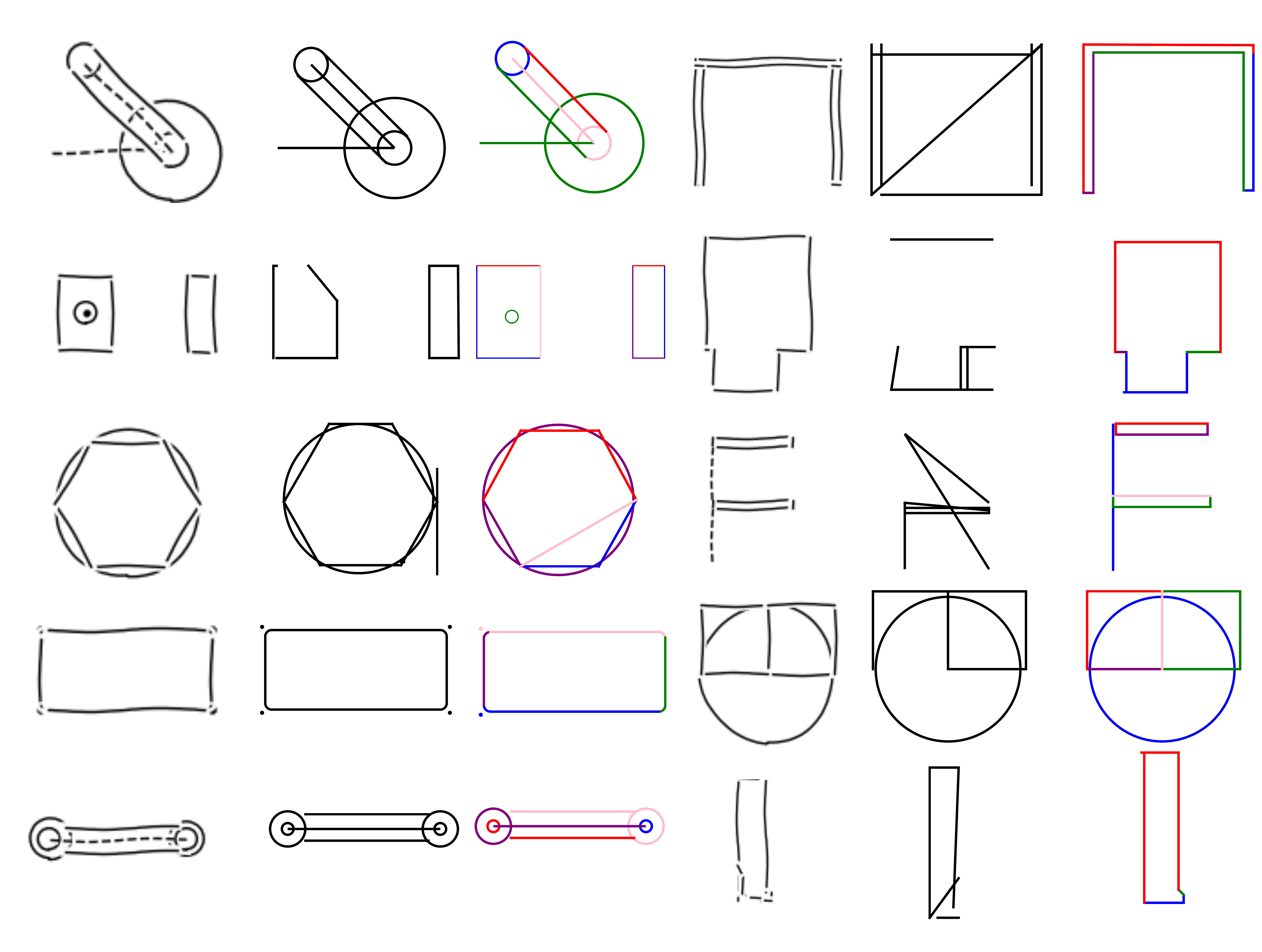}
    \caption{\textbf{Image conditioned generation}. Each example shows the input sketch image, the reconstruction result by autoregressive baseline, and our result.}
    \label{fig:image_conditioned}
\end{figure}

We provide quantitative evaluation in Table~\ref{tab:image_conditioned} and visual comparison in Fig.~\ref{fig:image_conditioned}, both showing that our model has superior performance than the autoregressive baseline, which again can be attributed to the more regular generation through sketch concept composition.

\begin{table}
\centering

\begin{tabular}{ccc}
        \toprule
        Config & Primitive  & Constraint \\
        \midrule
        Autoregressive & 0.575 &  0.301 \\
        Ours & \textbf{0.711} & \textbf{0.368}  \\
        \bottomrule
    \end{tabular}
    \vspace{1mm}
    \caption{\textbf{Image-conditioned generation.} F-scores are reported for primitives and constraints.}
    \label{tab:image_conditioned}
\end{table}

\subsection{Experiment on ``CAD as Language'' dataset}
\label{subsec:cadl_result}

We also conduct preliminary experiments of our method on the dataset of \cite{CADAsLanguage2021}, which comprises of millions of CAD sketches retrieved from OnShape \cite{onshape}; in comparison, the SketchGraphs dataset \cite{SketchGraphs} on which we have done the other experiments is similarly collected from OnShape but has a smaller scale.
We filter the dataset by removing trivial or semantically ambiguous sketches and confine the sketch complexity such that the total number of primitives and constraints is within $[20,90]$. In the end, we obtain about 2.5 million sketches and use 2.3 millions for training and the rest for testing. In comparison, there are about 1 million samples from SketchGraphs dataset used in the other experiments, where the maximum sketch graph size is 50 (Sec.~\ref{sec:results}). 

\begin{figure}
    \centering
    \includegraphics[width=1.0\linewidth]{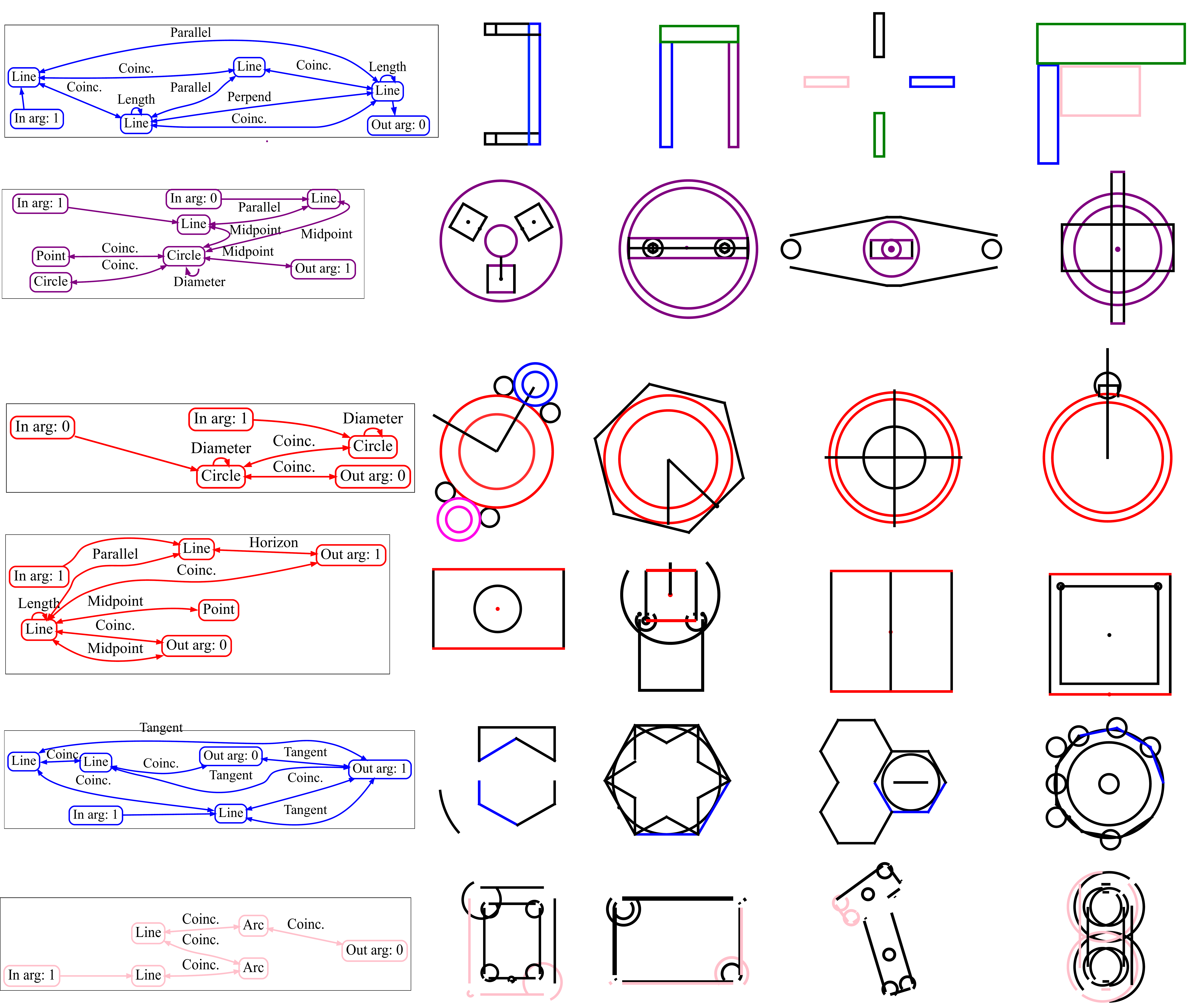}
    \caption{\textbf{ Examples of library concepts learned from the CAD as Language dataset \cite{CADAsLanguage2021} and their corresponding sketch instances}. Different instances of the same library are highlighted in different colors.}
    \label{fig:cadl_library}
\end{figure}

\begin{figure}
    \centering
    \includegraphics[width=1.0\linewidth]{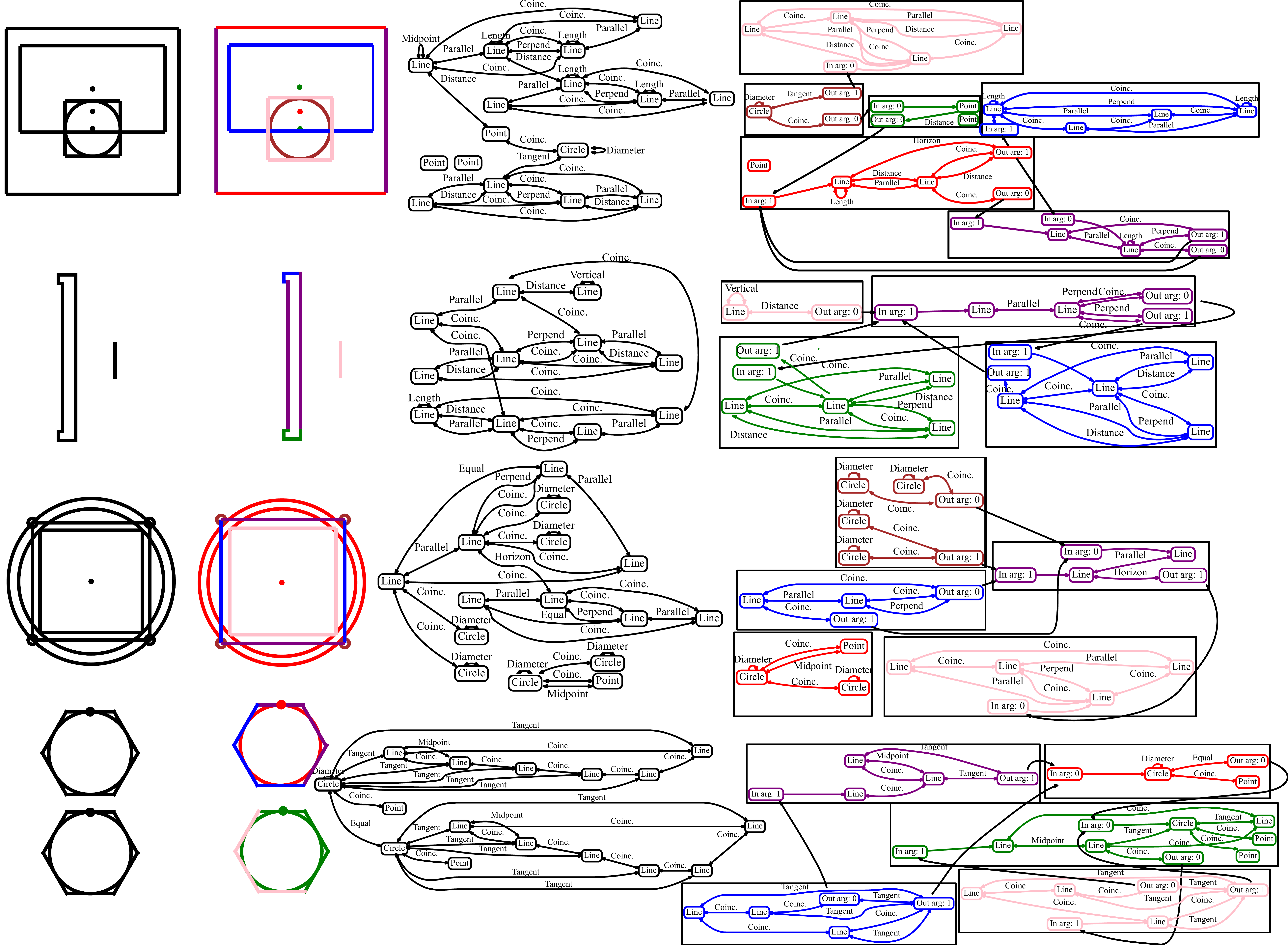}
    \caption{\textbf{ Design intent parsing learned on the CAD as Language dataset \cite{CADAsLanguage2021}}. Each example shows the input raw sketch and corresponding constraint graph (in black), as well as our interpreted sketch made of modular concepts and corresponding modular constraint graph, where primitives and constraints are colored according to their encapsulating concepts.}
    \label{fig:cadl_instance}
\end{figure}

To accommodate the increased complexity of this dataset, we increase the query number $k_{qry}$ to 6, the $\mathbb{L}^1$ library size to 15, and leave the rest hyperparameters unchanged. Examples of learned libraries and corresponding sketches are presented in Fig.~\ref{fig:cadl_library}. Examples of design intent parsing results are given in Fig.~\ref{fig:cadl_instance}.
We can see that our method obtains new modular concepts and parses more complex sketches; these results show that our framework works similarly on this new dataset.

\subsection{Broader impact}
\label{subsec:broad_impact}

This work potentially improves the efficiency of CAD sketch design, which however does not replace other critical procedures of CAD. For example, the discovered concepts do not necessarily meet structure safety constraints, and should be subject to checking and validation procedures according to specific applications.
The general methodology of program library induction presented in this work facilitates more structured and interpretable machine learning, which may enhance human and AI interaction but has no direct negative social impacts.

\end{document}